%% file: main.tex
\documentclass[runningheads]{llncs}

\usepackage{eccv}

\usepackage{eccvabbrv}

\usepackage{graphicx}
\usepackage{booktabs}

\usepackage[accsupp]{axessibility}  
\usepackage{booktabs}
\usepackage{multirow}
\usepackage{adjustbox}

\usepackage{algorithm}
\usepackage{algorithmic}

\usepackage{hyperref}

\usepackage{orcidlink}

\begin{document}

\title{Geodesic Flow Matching on a Riemannian Degradation Manifold for Blind Image Restoration} 

\titlerunning{Geodesic FM on Riemannian Degradation Manifold for BIR}

\author{Akshay Janardan Bankar\orcidlink{0009-0000-3577-1399} \and
Ankita Chatterjee \and
Sayan Banerjee \and
Shreyas Pandith \and
Kalakonda Sai Shashank \and
Amit Satish Unde
}

\authorrunning{A. Bankar, A. Chatterjee et al.}

\institute{Samsung Research Institute, Bangalore, India
}

\maketitle

\begin{abstract}
  Blind image restoration requires recovering clean images from observations corrupted by unknown and potentially mixed degradations. While recent deterministic flow-based methods model restoration as transport processes that map degraded images to clean ones, they typically rely on Euclidean interpolation, implicitly assuming linear degradation geometry. In this paper, we explicitly model degradations as points on a low-dimensional Riemannian manifold and formulate restoration as geodesic transport on the joint image–manifold space. Using a geodesic flow matching objective, we learn intrinsic transport dynamics that respect the curvature of degradation space. This framework generalizes linear flow matching, provides a principled treatment of mixed degradations as geodesic compositions, and yields a clean theoretical interpretation for generalization beyond observed degradations. 
\end{abstract}

\input{sec/introduction}

\input{sec/related_work}

\input{sec/method}

\input{sec/experiments}

\input{sec/conclusion}

\bibliographystyle{splncs04}
\bibliography{main}

\input{sec/supplementary}

\end{document}

%% file: sec/introduction.tex
\section{Introduction}

Blind image restoration seeks to recover high-quality images from observations corrupted by unknown degradations such as blur, haze, rain, or snow. In contrast to supervised settings with known corruption models, blind restoration must infer both the underlying image content and the degradation process, making the problem fundamentally ill-posed: distinct degradations can yield similar observations and mixed degradations can interact nonlinearly.

Generative approaches---notably diffusion models and continuous flow-based methods---have recently advanced restoration quality. Diffusion methods formulate restoration as a stochastic reverse process but often require many inference steps. Deterministic alternatives such as flow matching learn a vector field that transports degraded representations toward clean ones, enabling efficient inference with far fewer steps. Despite their effectiveness, degradations are typically handled implicitly, via unstructured conditioning or Euclidean embeddings, which ignores potential structure in degradation space and can limit stability and interpretability under diverse or severe corruptions.

In practice, degradations vary along meaningful axes such as severity and composition, suggesting that degradation states may lie on a low-dimensional \emph{manifold} rather than an unconstrained Euclidean space. Motivated by this observation, we investigate whether geometric inductive biases for degradation representations can improve blind restoration. We model degradation states as points on a \emph{Riemannian manifold} and formulate restoration as transport in a joint latent--manifold product space:
\[
(z_t, m_t) \in \mathcal{Z} \times \mathcal{M},
\]
where $\mathcal{Z}$ is the latent space of a pretrained autoencoder and $\mathcal{M}$ is a degradation manifold equipped with a Riemannian metric. Given a degraded image, a degradation encoder predicts a manifold state; the degradation state evolves along a geodesic toward a clean anchor while the latent representation follows a flow-matching path toward the clean latent. This coupling preserves compatibility with modern latent generative backbones while enabling structured degradation conditioning.

To obtain stable training and meaningful geometry, we adopt a two-stage strategy. Stage~1 learns a geometry-constrained degradation representation jointly with latent flow matching. Stage~2 freezes the degradation encoder and learns a manifold vector field that is explicitly aligned with the tangent bundle, enabling principled supervision of degradation dynamics and enforcing intrinsic (on-manifold) evolution. We instantiate the framework with Euclidean, spherical, and hyperbolic geometries and evaluate their effect on restoration quality and geometric behavior. Across multiple blind restoration tasks, non-Euclidean geometries provide useful inductive biases that stabilize degradation representation learning and yield more structured degradation trajectories; hyperbolic geometry is particularly effective under severe degradations while maintaining strong restoration quality.

\paragraph{Contributions.}
Our main contributions are:
\begin{itemize}
\item We propose a \textbf{Riemannian degradation manifold} formulation for blind image restoration, representing degradations as points on a structured geometric space rather than unconstrained embeddings.
\item We introduce a \textbf{product-space transport} framework that couples latent flow matching with geodesic evolution of degradation states.
\item We develop a \textbf{two-stage training strategy} that learns geometry-constrained degradation representations and tangent-bundle-aligned manifold dynamics.
\item Through experiments on multiple restoration tasks, we show that \textbf{non-Euclidean geometries improve representation stability and yield structured degradation states} while maintaining competitive restoration performance.
\end{itemize}

%% file: sec/related_work.tex
\section{Related Work}
\label{sec:related_work}

\paragraph{Flow matching and bridge-based generative modeling.}
Flow Matching (FM) trains continuous normalizing flows with a simulation-free objective by regressing a vector field tied to a chosen conditional probability path \cite{lipman2023flowmatching}. It has been generalized to manifolds via \emph{Riemannian Flow Matching} (RFM), enabling principled vector-field learning on common geometries with closed-form targets \cite{chen2023rfm}. In parallel, bridge-based and ``interpolant'' views connect diffusion/score models and deterministic flows through path-wise objectives; latent-space bridges such as Latent Bridge Matching (LBM) show that learning transport in VAE latents can enable fast image-to-image translation in only a few steps \cite{chadebec2025lbm}. While these works motivate transport in compact latent representations, they do not model degradations as states on an explicit Riemannian manifold with tangent-bundle alignment.

\paragraph{Generative image restoration with flows and diffusion.}
Diffusion and score-based restoration typically follow stochastic reverse-time sampling and often require many steps, motivating continuous-time flow alternatives for efficiency. ResFlow \cite{qin2025resflow}, for example, models degradation as a deterministic continuous flow with an auxiliary process to resolve ambiguity, learning via velocity matching and achieving strong results in few steps \cite{qin2025resflow}. Our approach is complementary: instead of adding an auxiliary Euclidean process, we introduce an explicit \emph{degradation manifold} and learn restoration dynamics in a \emph{product space} coupling latent content transport with manifold-structured degradation states.

\paragraph{Latent diffusion backbones.}
Latent Diffusion Models (LDMs), popularized by Stable Diffusion, run diffusion in a learned latent space and decode to pixels with a pretrained autoencoder, enabling high-quality generation at lower cost \cite{rombach2022ldm}. For restoration, such latent backbones provide strong priors and effective conditioning (e.g., cross-attention tokens). We build on a Stable Diffusion latent backbone while adding a geometrically structured degradation representation that improves training stability and interpretability.

\paragraph{Non-Euclidean representation learning (spherical and hyperbolic).}
Non-Euclidean embeddings are widely used for structured representations. Hyperbolic embeddings in the Poincar\'e ball \cite{nickel2017poincare} and Lorentz model \cite{nickel2018lorentz} offer expanding distances, and hyperbolic neural networks extend common operations (MLPs, attention) to curved spaces \cite{ganea2018hyperbolic}. These geometries can regularize and organize low-dimensional embeddings. Here, we adopt spherical and hyperbolic manifolds as inductive biases for degradation states and analyze their effects on restoration performance and optimization stability.

%% file: sec/method.tex
\section{Method}
\label{sec:method}

\subsection{Overview and Motivation}
\label{sec:method_overview}

Blind image restoration is inherently ambiguous: multiple degradation mechanisms can map different clean images to similar degraded observations. Flow-based restoration methods learn a transport map from degraded to clean domains by predicting a velocity field, but in standard formulations the degradation information is either implicit in the input or encoded in unconstrained Euclidean embeddings \cite{qin2025resflow}. We propose to explicitly \emph{factor} our proposed flow based restoration process into (i) content transport in a latent space and (ii) a structured degradation state evolving on a Riemannian manifold. This yields a product-space formulation:
\begin{equation}
(z_t, m_t) \in \mathcal{Z} \times \mathcal{M},
\end{equation}
where $\mathcal{Z}$ is the latent space of a Stable Diffusion autoencoder \cite{rombach2022ldm}, and $\mathcal{M}$ is a learned degradation manifold equipped with a Riemannian metric. Intuitively, $z_t$ models the image content along the restoration trajectory, while $m_t$ provides an explicit, geometry-constrained coordinate system for degradation.

\begin{figure*}[ht]
    \centering
    \includegraphics[width=0.4\linewidth]{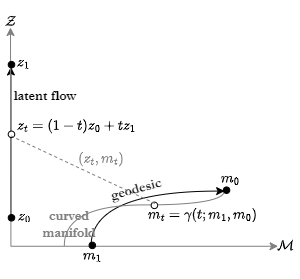}
    \caption{\textbf{Joint latent--manifold product space for restoration.} We formulate blind restoration as transport in $\mathcal{Z}\times\mathcal{M}$, where $\mathcal{Z}$ is the Stable Diffusion latent space and $\mathcal{M}$ a learned degradation manifold. For a degraded/clean pair, latents follow a flow-matching path $z_t$ from $z_0$ (degraded) to $z_1$ (clean), while degradation states traverse a geodesic $m_t=\gamma(t;m_1,m_0)$ from the inferred embedding $m_1$ to the clean anchor $m_0$. Conditioning the restoration UNet on $m_t$ couples content transport with geometry-constrained degradation evolution.}
\label{fig:joint_product_space}
    \label{fig:product_manifold}
\end{figure*}

We train in two stages for stability and identifiability:
\begin{itemize}
    \item \textbf{Stage-1 (Representation Learning):} learn a geometry-constrained degradation representation and condition latent flow matching on it.
    \item \textbf{Stage-2 (Tangent-Bundle Learning):} freeze the degradation representation and learn a manifold vector field aligned with the tangent bundle.
\end{itemize}
This separation avoids moving-target supervision for manifold velocities and yields a cleaner geometric interpretation.

\begin{figure*}[t]
    \centering
    \includegraphics[width=0.9\linewidth]{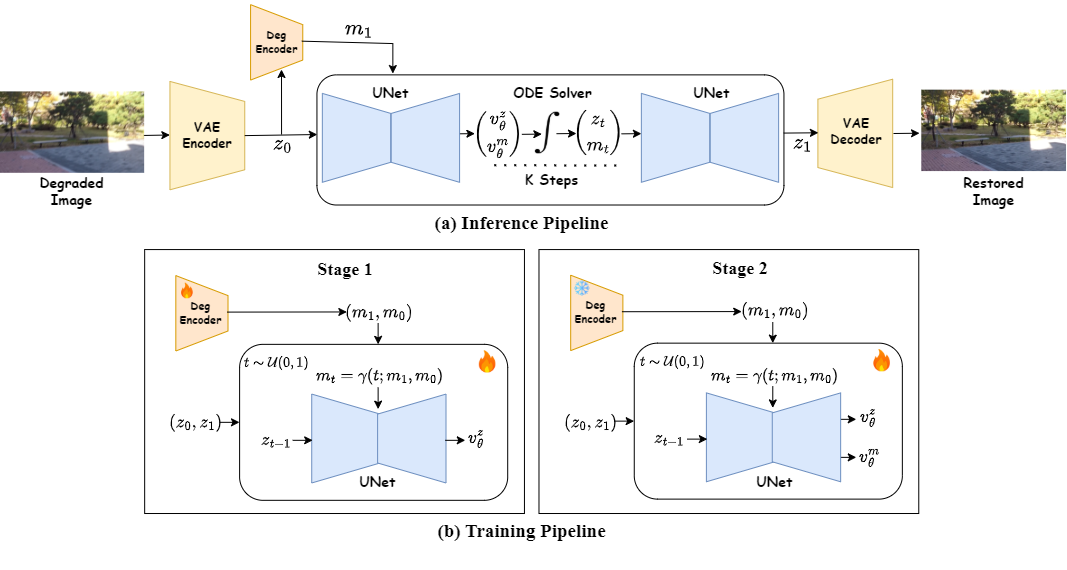}
    \caption{\textbf{Inference and two-stage training pipeline.} \textbf{(a) Inference.} A degraded image is encoded to $z_0$ with a frozen Stable Diffusion autoencoder. A degradation encoder predicts $m_1$, defining a time-varying manifold state $m_t=\gamma(t;m_1,m_0)$. Conditioned on $(t,m_t)$, the UNet predicts the latent vector field $v_\theta^z(z_t,t,m_t)$ and integrates it for $K$ steps (e.g., Heun/RK2) to obtain the restored latent, which is decoded to the output image. \textbf{(b) Training.} \textbf{Stage 1} learns geometry-aware degradation embeddings and the latent restoration flow via flow-matching supervision with $t\sim\mathcal{U}(0,1)$. \textbf{Stage 2} freezes the degradation encoder and trains a manifold head to predict a tangent-aligned field $v_\theta^m(m_t,t,e)$ while continuing the latent flow objective, keeping manifold dynamics consistent with $\mathcal{M}$.}
    \label{fig:Training and inference pipeline main figure}
\end{figure*}

\subsection{Latent Flow Matching for Restoration}
\label{sec:latent_fm}

Given a degraded image $y$ and its clean counterpart $x$, we encode them into latent space using a pretrained autoencoder:
\begin{equation}
z_0 = E_\phi(y), \qquad z_1 = E_\phi(x),
\end{equation}
where $E_\phi$ is the (frozen) Stable Diffusion encoder \cite{rombach2022ldm}. We adopt a standard flow matching interpolation path in latent space \cite{lipman2023flowmatching}:
\begin{equation}
z_t = (1-t) z_0 + t z_1, \qquad t \sim \mathcal{U}(0,1),
\label{eq:zt_linear}
\end{equation}
with target velocity
\begin{equation}
v_z^\star = \frac{d}{dt} z_t = z_1 - z_0.
\label{eq:vz_star}
\end{equation}
A UNet backbone $f_\theta$ predicts a latent velocity field $v_\theta^z(z_t,t,\mathrm{cond})$, trained with the flow matching regression objective:
\begin{equation}
\mathcal{L}_z = 
\mathbb{E}_{t,z_0,z_1}
\left[\,
\left\|v_\theta^z(z_t,t,\mathrm{cond}) - v_z^\star \right\|_2^2
\right].
\label{eq:Lz}
\end{equation}
In standard latent FM, $\mathrm{cond}$ is either absent or unconstrained. Our key contribution is to construct $\mathrm{cond}$ from a structured degradation state $m_t \in \mathcal{M}$ and to explicitly learn dynamics on $\mathcal{M}$ in Stage-2.

\subsection{Stage-1: Learning a Geometry-Constrained Degradation State}
\label{sec:stage1}

\paragraph{Degradation encoder.}
We introduce a degradation encoder $g_\psi$ that maps degraded latents to a point on a Riemannian manifold:
\begin{equation}
m_1 = g_\psi(z_0) \in \mathcal{M}, \qquad e = h_\psi(z_0),
\label{eq:deg_enc}
\end{equation}
where $e$ is an optional scalar/vector severity descriptor used as auxiliary conditioning. We define a fixed reference point $m_0 \in \mathcal{M}$ representing the clean anchor. During Stage-1, $g_\psi$ is \emph{trainable} and is optimized only through the restoration objective (Eq.~\ref{eq:Lz}), i.e., the geometry is learned as an inductive bias that improves restoration.

\paragraph{Geodesic interpolation on the manifold.}
Given $(m_1, m_0)$, we define an intrinsic interpolation path via the geodesic $\gamma$:
\begin{equation}
m_t = \gamma(t; m_1, m_0) \in \mathcal{M}.
\label{eq:mt_geo}
\end{equation}
For Euclidean space $\mathbb{R}^d$ this reduces to linear interpolation; for spherical and hyperbolic geometries, it is a curve. The role of $m_t$ is to provide a \emph{time-consistent} degradation state that can be injected into the restoration network.

\paragraph{Conditioning via cross-attention.}
We condition the UNet using cross-attention tokens. Specifically, we project the manifold state to the UNet context dimension and concatenate it with the empty-text token:
\begin{equation}
\mathbf{c}_t = \left[\mathrm{Emb}(\varnothing),\; W_m m_t\right],
\label{eq:context}
\end{equation}
and set $\mathrm{cond}\equiv \mathbf{c}_t$ in Eq.~\ref{eq:Lz}. This design preserves the generative prior of the Stable Diffusion backbone while injecting degradation-specific information.

\paragraph{Stage-1 objective.}
Stage-1 optimizes:
\begin{equation}
\min_{\theta,\psi}\; \mathcal{L}_z,
\label{eq:stage1_obj}
\end{equation}
with the VAE frozen. Empirically, we observe that non-Euclidean geometries (sphere/hyperbolic) improve optimization stability by constraining representation drift and inducing smoother severity ordering (Sec.~\ref{sec:experiments}).

\subsection{Stage-2: Tangent-Bundle Aligned Manifold Vector Fields}
\label{sec:stage2}

Stage-1 produces a meaningful degradation coordinate system but does not constrain the \emph{dynamics} on $\mathcal{M}$. In Stage-2, we freeze $g_\psi$ to fix the degradation representation and learn an explicit manifold velocity field. This yields two benefits: (i) stable supervision (no moving target), and (ii) a principled geometric claim: the learned field is a section of the tangent bundle.

\paragraph{Geodesic velocity target.}
Given the geodesic $m_t=\gamma(t;m_1,m_0)$, we compute its derivative
\begin{equation}
v_m^\star(t) = \dot m_t = \frac{d}{dt}\gamma(t;m_1,m_0),
\label{eq:vm_star}
\end{equation}
which lies in the tangent space $T_{m_t}\mathcal{M}$. For spherical/hyperbolic geometries, $\dot m_t$ is computed using closed-form expressions consistent with the exponential-map parameterization (please refer to Sec.~\ref{sec:geometry_ops}).

\paragraph{Manifold velocity prediction and tangent projection.}
We extend the UNet with a small head to predict a manifold velocity $v_\theta^m$. To ensure geometric validity, we project predictions onto the tangent space:
\begin{equation}
\tilde v_\theta^m = \Pi_{T_{m_t}\mathcal{M}}\!\left(v_\theta^m\right),
\label{eq:tangent_proj}
\end{equation}
where $\Pi_{T_{m_t}\mathcal{M}}$ is the geometry-specific tangent projection operator (Sec.~\ref{sec:geometry_ops}). We then supervise in the intrinsic metric:
\begin{equation}
\mathcal{L}_m =
\mathbb{E}\left[
\left\| \tilde v_\theta^m - v_m^\star \right\|_{g(m_t)}^2
\right].
\label{eq:Lm}
\end{equation}
where $\lvert\rvert\cdot\lvert\rvert_{g(m_t)}$ is the Riemannian metric norm defined over the tangent space $T_{m_t}\mathcal{M}$
\paragraph{Stage-2 objective.}
Stage-2 optimizes UNet parameters while freezing the degradation encoder:
\begin{equation}
\min_{\theta}\; \mathcal{L}_z + \lambda_m \mathcal{L}_m,
\label{eq:stage2_obj}
\end{equation}
where $\lambda_m$ balances restoration and geometric alignment. In practice, we keep $\mathcal{L}_z$ active to maintain restoration accuracy while learning manifold-consistent dynamics.

\subsection{Inference}
\label{sec:inference}

At test time, given a degraded image $y$, we encode to latent space $z_0=E_\phi(y)$ and obtain a degradation state via the (frozen) encoder:
\begin{equation}
m_1 = g_\psi(z_0), \qquad m_t = \gamma(t;m_1,m_0).
\end{equation}
We then integrate the latent ODE induced by the learned vector field:
\begin{equation}
\frac{d z}{dt} = v_\theta^z(z(t), t, \mathbf{c}_t),
\label{eq:ode_latent}
\end{equation}
from $t=0$ to $t=1$ (or the equivalent direction depending on the chosen convention), using a fixed-step solver (e.g., Euler or Heun). The restored image is obtained by decoding $\hat x = D_\phi(\hat z_1)$ using the frozen decoder $D_\phi$ \cite{rombach2022ldm}. In our main implementation, $m_t$ is used as conditioning along the trajectory; optionally, $m_t$ can also be updated via $\dot m = \tilde v_\theta^m$ using the manifold exponential map, though we find conditioning-only inference to be sufficient and more stable.

\subsection{Geometry Operations (Exp/Log/Projection)}
\label{sec:geometry_ops}

We summarize the geometry-specific operators required by our framework: (i) manifold projection to enforce $m\in\mathcal{M}$, (ii) tangent projection to enforce $v\in T_m\mathcal{M}$, and (iii) geodesic interpolation $m_t=\gamma(t;m_1,m_0)$ implemented via exponential and logarithmic maps. Complete closed-form expressions and numerical stabilizations are provided in supplementary.

\paragraph{Euclidean.}
For $\mathcal{M}=\mathbb{R}^d$, $\gamma(t)=(1-t)m_1+t m_0$ and no projection is required.

\paragraph{Sphere.}
For $\mathcal{M}=\mathbb{S}^{d-1}$, we enforce $\|m\|_2=1$ via normalization, and project velocities onto the tangent space:
\begin{equation}
\Pi_{T_m\mathbb{S}^{d-1}}(v) = v - (m^\top v)\,m.
\end{equation}
Geodesic interpolation is given by $m_t=\mathrm{Exp}_{m_1}(t\,\mathrm{Log}_{m_1}(m_0))$.

\paragraph{Hyperboloid (Lorentz).}
For $\mathcal{M}=\mathbb{H}^d_c$ with curvature $-c$, we enforce $\langle m,m\rangle_L=-1/c$ and $m_0>0$. Tangent projection is:
\begin{equation}
\Pi_{T_m\mathbb{H}^d_c}(v) = v + c\,\langle m,v\rangle_L\,m,
\end{equation}
where, $\langle x,y\rangle_L=-x_0y_0+\sum_{i=1}^d x_i y_i$. Geodesic interpolation uses $m_t=\mathrm{Exp}_{m_1}$\\$(t\,\mathrm{Log}_{m_1}(m_0))$ in the Lorentz model (detailed formulation presented in supplementary).

%% file: sec/experiments.tex
\section{Experiments}
\label{sec:experiments}

We evaluate our method from two complementary perspectives: (i) restoration performance, measured through distortion- and distribution-based metrics, and (ii) geometric validity of the learned degradation state, measured through manifold diagnostics.

\subsection{Experimental Setup}
\label{sec:exp_setup}

\paragraph{Tasks.}
We consider four blind image restoration tasks: deblurring, dehazing, deraining, and desnowing. Each task is trained independently to isolate the effect of the degradation geometry and avoid cross-task confounds.

\paragraph{Backbone and conditioning.}
We operate in the latent space of a pretrained Stable Diffusion autoencoder \cite{rombach2022ldm}. The restoration vector field is parameterized by a UNet backbone. The degradation state $m_t$ is injected through cross-attention by concatenating a projected manifold token with the empty-text embedding, i.e., $\mathbf{c}_t=[\mathrm{Emb}(\varnothing),\,W_m m_t]$.

\paragraph{Geometries.}
We compare three manifold instantiations for the degradation state: Euclidean ($\mathbb{R}^d$), spherical ($\mathbb{S}^{d-1}$), and hyperbolic (Lorentz hyperboloid $\mathbb{H}^d_c$). Geometry-specific operations are summarized in Sec.~\ref{sec:geometry_ops}, with full closed-form expressions provided in Supplementary..

\paragraph{Two-stage protocol.}
Stage-1 jointly trains the degradation encoder and latent flow matching objective. Stage-2 freezes the degradation encoder and adds tangent-bundle aligned manifold supervision while keeping the latent restoration objective active. This design avoids moving-target instability for manifold velocity supervision.

\paragraph{Metrics.}
For restoration fidelity we report PSNR$\uparrow$, SSIM$\uparrow$, and LPIPS$\downarrow$. For distributional/perceptual quality, evaluated on separate task-specific benchmarks, we report FID$\downarrow$ \cite{heusel2017fid} and KID$\downarrow$ \cite{binkowski2018kid}. For geometric analysis, we report manifold constraint residual, tangent violation magnitude, severity--distance correlation, and related qualitative manifold diagnostics.






\subsection{Datasets and Evaluation Protocols}
\label{sec:datasets}

We evaluate four restoration tasks (deblurring, dehazing, deraining, and desnowing). For each task, we report fidelity metrics (PSNR/SSIM/LPIPS) on a paired benchmark and perceptual distribution metrics (FID/KID) on a complementary real-image benchmark, following common diffusion-style evaluation practice. Specifically, we use DPDD \cite{abdelhamed2019dpdd} and RealBlur-J \cite{rim2020realblur} for deblurring; NH-HAZE \cite{ancuti2020nhhaze} and RESIDE \cite{li2018reside} for dehazing; LHP \cite{guo2023lhprain} and RealRain/RealTest \cite{yang2020realrain} for deraining; and RealSnow \cite{liu2021realsnow} together with the ``Realistic'' benchmark from Dense-Snow \cite{chen2021densely} for desnowing.

\subsection{Implementation Details}
\label{sec:implementation}

During training we sample $t\sim\mathcal{U}(0,1)$ and regress the flow-matching target velocity. At inference we integrate the latent ODE using a small number of fixed step. For spherical and hyperbolic geometries, we use clamped $\mathrm{Exp}/\mathrm{Log}$ maps and explicit tangent projections.

\subsection{Main Restoration Results}
\label{sec:main_results}

\paragraph{Quantitative results.}
Table~\ref{tab:deblur_dpdd} reports deblurring performance on DPDD \cite{abdelhamed2019dpdd} against strong CNN-, transformer-, and flow-based baselines. For real-world tasks, Table~\ref{tab:real_resflow_style} reports PSNR/SSIM comparisons on NH-HAZE, LHP, and RealSnow using a compact task-wise format. For distributional quality, Table~\ref{tab:diffusion_fid} reports FID/KID comparisons on RealBlur-J, RESIDE, RealRain, and Dense-Snow benchmarks.

Across these benchmarks, our final hyperbolic model achieves competitive or improved restoration performance. The gains are particularly notable on challenging deblurring cases and remain strong on real-world degradations, indicating that the proposed degradation manifold improves restoration fidelity while remaining competitive with high-capacity restoration architectures.

\begin{table*}[t]
\centering
\caption{Single-image defocus deblurring comparison on the DPDD~\cite{abdelhamed2019dpdd} dataset.
Best and second-best results are highlighted in \textbf{bold} and \underline{underline}.}
\label{tab:deblur_dpdd}
\scriptsize
\setlength{\tabcolsep}{2.6pt}
\renewcommand{\arraystretch}{0.95}
\begin{adjustbox}{max width=\linewidth}
\begin{tabular}{lcccccccccccc}
\toprule
\multirow{2}{*}{Method}
& \multicolumn{4}{c}{Indoor}
& \multicolumn{4}{c}{Outdoor}
& \multicolumn{4}{c}{Combined} \\
\cmidrule(lr){2-5} \cmidrule(lr){6-9} \cmidrule(lr){10-13}
& PSNR$\uparrow$ & SSIM$\uparrow$ & MAE$\downarrow$ & LPIPS$\downarrow$
& PSNR$\uparrow$ & SSIM$\uparrow$ & MAE$\downarrow$ & LPIPS$\downarrow$
& PSNR$\uparrow$ & SSIM$\uparrow$ & MAE$\downarrow$ & LPIPS$\downarrow$ \\
\midrule

\multicolumn{13}{c}{\textit{CNN / Transformer-based}} \\
Restormer~\cite{zamir2022restormer} & 28.87 & 0.882 & 0.025 & 0.145 & 23.24 & 0.743 & 0.050 & 0.209 & 25.90 & 0.811 & 0.038 & 0.178 \\
FocalNet~\cite{cui2023focal} & 29.10 & 0.876 & 0.024 & 0.173 & 23.14 & 0.743 & 0.049 & 0.246 & 26.18 & 0.808 & 0.037 & 0.153 \\
DTPM-4~\cite{ye2024learning} & -- & -- & -- & -- & -- & -- & -- & -- & 25.98 & 0.823 & 0.038 & 0.153 \\

\multicolumn{13}{c}{\textit{Flow-based}} \\
ResFlow~\cite{Qin2025} & \textbf{29.81} & 0.907 & \underline{0.022} & \textbf{0.096} & \textbf{24.25} & \textbf{0.782} & \underline{0.046} & \textbf{0.166} & \textbf{26.96} & \textbf{0.842} & \underline{0.034} & \underline{0.131} \\

\midrule
\multicolumn{13}{c}{\textit{Ours}} \\
Euclidean (Stage-2) & 28.57 & 0.886 & 0.024 & 0.128 & 22.61 & 0.743 & 0.049 & 0.198 & 25.59 & 0.814 & 0.036 & 0.171 \\
Sphere (Stage-2) & 29.01 & \underline{0.909} & 0.023 & 0.108 & 23.74 & 0.761 & 0.046 & 0.179 & 26.38 & 0.835 & \underline{0.034} & 0.153 \\
Hyperbolic (Stage-2) & \underline{29.23} & \textbf{0.912} & \textbf{0.020} & \underline{0.104} & \underline{23.89} & \underline{0.768} & \textbf{0.044} & \underline{0.174} & \underline{26.56} & \underline{0.840} & \textbf{0.032} & \textbf{0.130} \\
\bottomrule
\end{tabular}
\end{adjustbox}
\vspace{-1mm}
\end{table*}

\begin{table*}[t]
\centering
\caption{\textbf{Real-world datasets.}
Dehazing results on NH-HAZE~\cite{ancuti2020nhhaze},
Deraining results on LHP~\cite{guo2023lhprain},
and Desnowing results on RealSnow~\cite{liu2021realsnow}.
Higher is better.}
\label{tab:real_resflow_style}
\vspace{-2mm}
\scriptsize
\setlength{\tabcolsep}{3pt}
\renewcommand{\arraystretch}{1.05}
\begin{adjustbox}{max width=\linewidth}
\begin{tabular}{l cc | l cc | l cc}
\toprule
\multicolumn{1}{c}{\textbf{Method}} &
\multicolumn{2}{c}{\textbf{Dehazing}} &
\multicolumn{1}{c}{\textbf{Method}} &
\multicolumn{2}{c}{\textbf{Deraining}} &
\multicolumn{1}{c}{\textbf{Method}} &
\multicolumn{2}{c}{\textbf{Desnowing}} \\
\cmidrule(lr){2-3}\cmidrule(lr){5-6}\cmidrule(lr){8-9}
 & PSNR & SSIM &  & PSNR & SSIM &  & PSNR & SSIM \\
\midrule

DeHamer~\cite{guo2022image} & 20.66 & 0.68 &
MPRNet~\cite{zamir2021mprnet} & 33.34 & 0.930 &
NAFNet~\cite{chen2022nafnet} & 31.44 & 0.919 \\

FocalNet~\cite{cui2023focal} & 20.43 & 0.79 &
SCD-Former~\cite{guo2023sky} & \textbf{34.33} & \underline{0.946} &
TransWeather~\cite{valanarasu2022transweather} & 31.13 & \underline{0.922} \\

PMNet~\cite{ye2022pmnet} & 20.42 & 0.73 &
Restormer~\cite{zamir2022restormer} & 29.97 & 0.921 &
Restormer~\cite{zamir2022restormer} & 30.52 & 0.909 \\

ResFlow\cite{Qin2025} & \textbf{21.44} & \textbf{0.79} &
ResFlow\cite{Qin2025} & 32.82 & 0.936 &
ResFlow\cite{Qin2025} & 31.86 & 0.917 \\

\midrule
\textbf{Ours (Euclidean)} & 20.59 & 0.71 &
\textbf{Ours (Euclidean)} & 31.44 & 0.920 &
\textbf{Ours (Euclidean)} & 31.91 & 0.894 \\

\textbf{Ours (Spherical)} & 21.20 & 0.73 &
\textbf{Ours (Spherical)} & 32.16 & 0.945 &
\textbf{Ours (Spherical)} & \underline{32.18} & 0.902 \\

\textbf{Ours (Hyperbolic)} & \underline{21.17} & \underline{0.74} &
\textbf{Ours (Hyperbolic)} & \underline{33.85} & \textbf{0.967} &
\textbf{Ours (Hyperbolic)} & \textbf{32.54} & \textbf{0.929} \\

\bottomrule
\end{tabular}
\end{adjustbox}
\vspace{-2mm}
\end{table*}

\begin{figure*}[ht]
    \centering
    \includegraphics[width=1.0\linewidth]{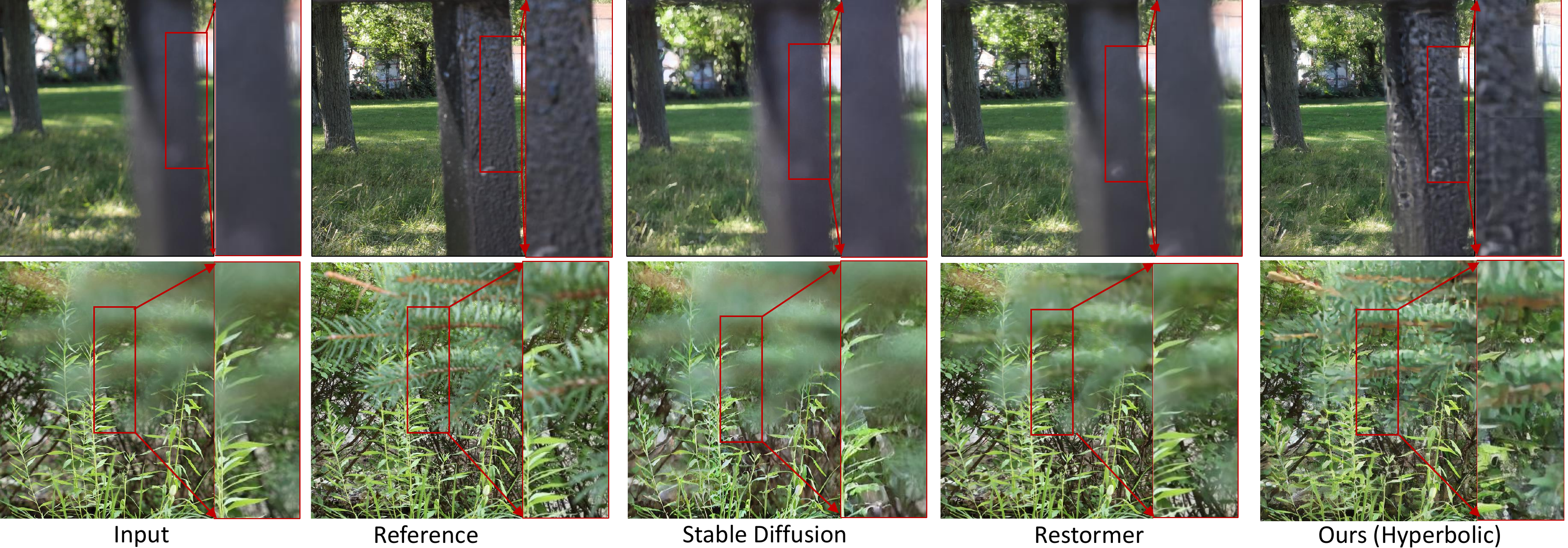}
    \caption{\textbf{Qualitative comparison on image deblurring.}
Visual comparison of our final \emph{hyperbolic} model against representative restoration baselines on challenging deblurring examples.
Our method recovers sharper edges, preserves fine structures more faithfully, and suppresses ringing and oversmoothing artifacts more effectively than competing methods.
The improvement is particularly visible in regions with severe blur, where the proposed geometry-aware latent flow produces clearer structural restoration and more natural image details.}
    \label{fig:qual_deblur}
\end{figure*}

\begin{figure*}[ht]
    \centering
    \includegraphics[width=1.0\linewidth]{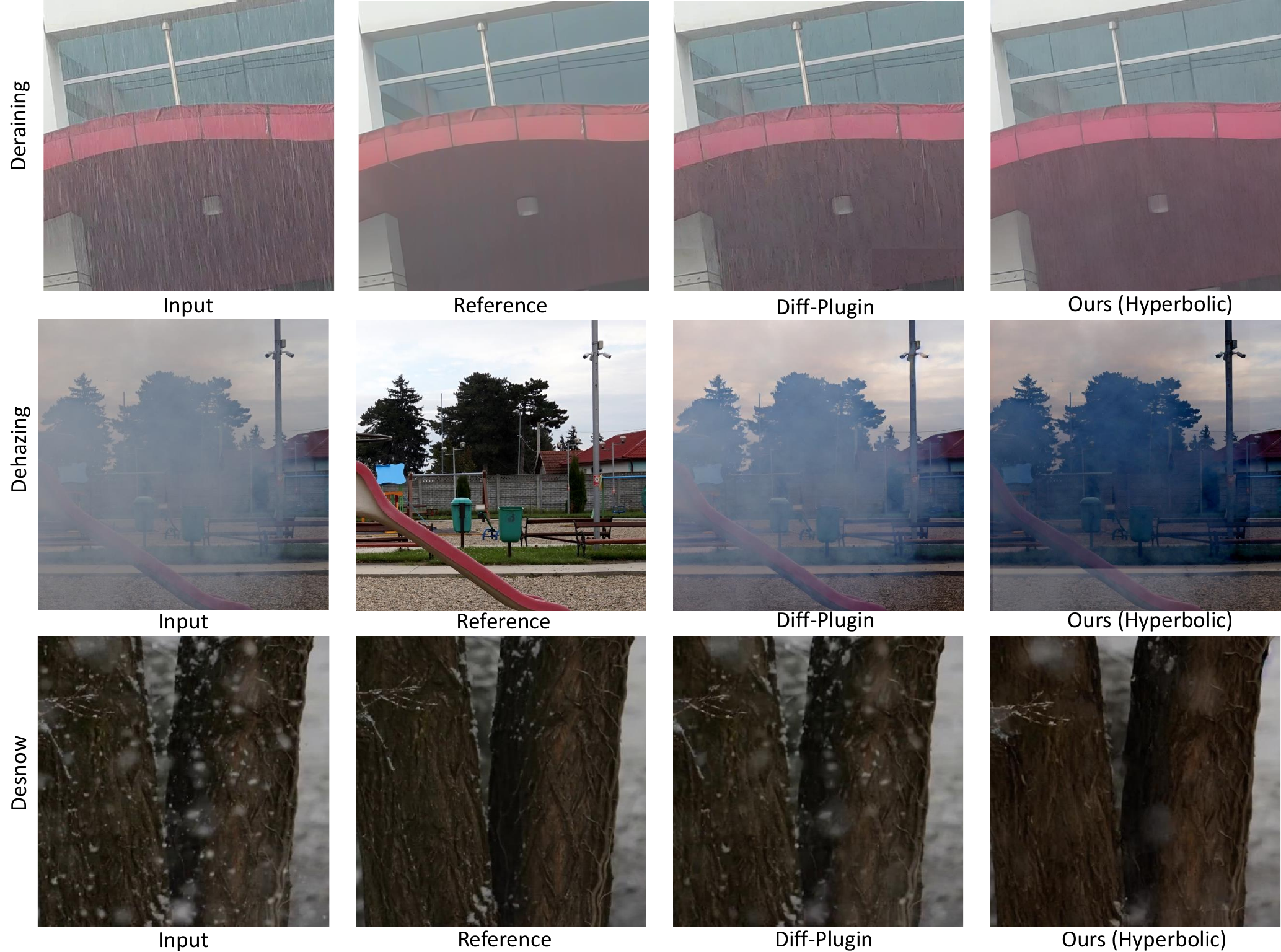}
    \caption{\textbf{Qualitative comparison on dehazing, deraining, and desnowing.} Visual results on three tasks beyond deblurring. Across all settings, our final \emph{hyperbolic} model produces cleaner, more faithful restorations—improving visibility and contrast under haze, removing dense rain streaks, and recovering snow-occluded details—with sharper structures and fewer residual weather artifacts than competing methods.}
    \label{fig:qual_weather}
\end{figure*}

\paragraph{Qualitative results.}
Figure~\ref{fig:qual_deblur} shows representative deblurring results, and Figure~\ref{fig:qual_weather} shows qualitative comparisons for dehazing, deraining, and desnowing. Compared with strong transformer-, flow-, and diffusion-based baselines, our final hyperbolic model restores sharper structures, preserves fine details more faithfully, and suppresses visible restoration artifacts more effectively, especially under severe degradations. These visual trends are consistent with the quantitative improvements reported in Tables~\ref{tab:deblur_dpdd}, \ref{tab:real_resflow_style}, and \ref{tab:diffusion_fid}.

\subsection{Manifold Diagnostics}
\label{sec:manifold_diagnostics}

We next examine whether the learned degradation state behaves consistently with the intended manifold structure.

\paragraph{Constraint satisfaction.}
For spherical and hyperbolic geometries, the manifold constraint residual~\ref{tab:stability} remains close to zero across samples and timesteps, indicating that the learned states stay on the target manifold under the chosen parameterization.

\paragraph{Severity ordering.}
We compute intrinsic geodesic distances and evaluate their correlation with degradation severity~\ref{tab:stability}. The learned manifold exhibits a consistent monotonic relationship between severity and intrinsic distance, showing that degradation strength is reflected in the geometry of the learned representation rather than an arbitrary embedding scale.

\begin{figure*}[ht]
    \centering
    \includegraphics[width=1.0\linewidth]{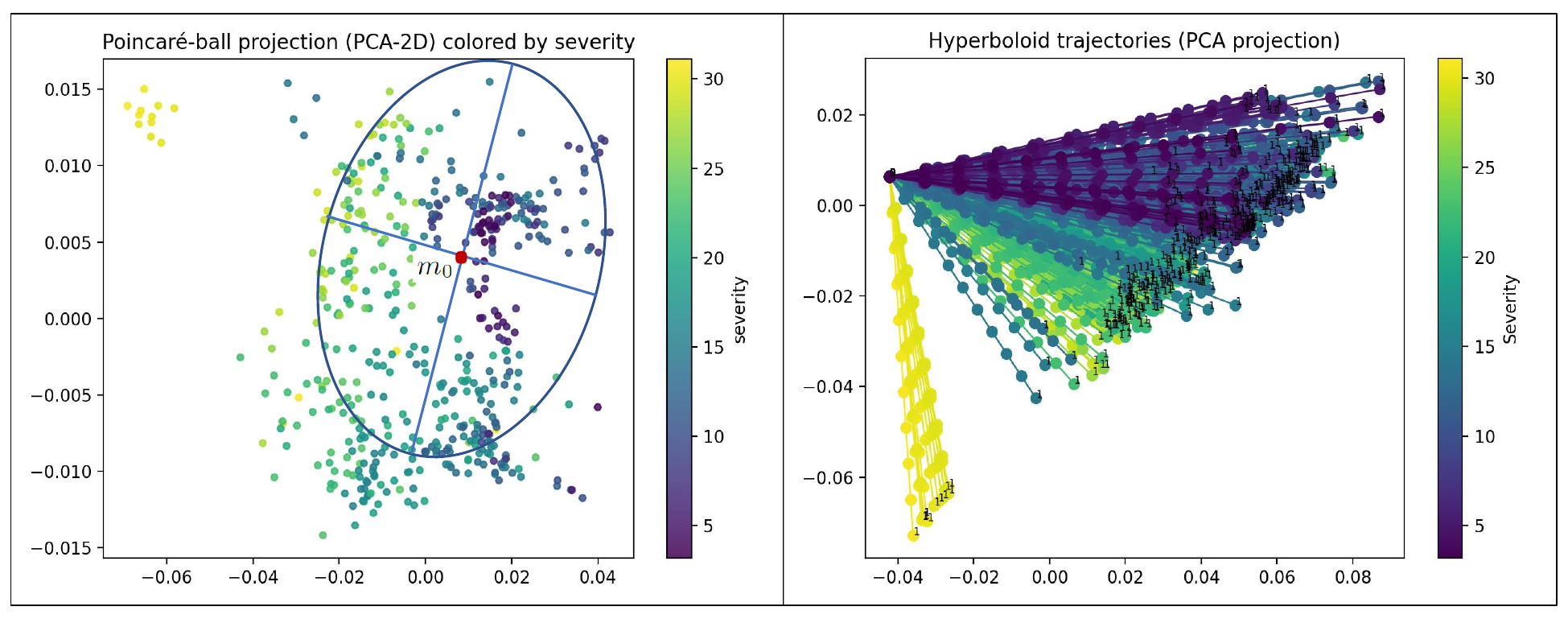}
\caption{\textbf{Learned hyperbolic degradation manifold.} (\textbf{Left}) Poincar'e-ball (PCA-projected) degradation embeddings colored by severity; the clean reference $m_0$ lies at the manifold center, and larger distance from $m_0$ indicates stronger degradation. (\textbf{Right}) Representative geodesic restoration paths $m_t=\gamma(t;m_1,m_0)$ that move degraded embeddings $m_1$ toward $m_0$, illustrating geometry-aware degradation evolution.}
    \label{fig:hyp_manifold}
\end{figure*}

\paragraph{Hyperbolic visualization.}
For hyperbolic geometry, we visualize learned embeddings using Poincar\'e projections~\ref{fig:hyp_manifold} together with iso-distance structure. Severe degradations occupy larger intrinsic radii, while cleaner states remain closer to the anchor point, yielding an interpretable progression of degradation severity.

\begin{table}[t]
\centering
\caption{Stability and geometric diagnostics comparison (averaged over tasks).}
\label{tab:stability}
\begin{adjustbox}{max width=\linewidth}
\begin{tabular}{lcccc}
\toprule
\textbf{Geometry} & \textbf{Constraint Residual} $\downarrow$ & \textbf{Tangent Violation} $\downarrow$ & \textbf{Speed Variance} $\downarrow$ & \textbf{Severity Corr.} $\uparrow$ \\
\midrule
Euclidean & $10{^{-3}}$ & -- & 0.716 & 0.67 \\
Sphere & $10{^{-3}}$ & -- & 0.624 & 0.75 \\
Hyperbolic & \textbf{$10{^{-4}}$} & \textbf{$10{^{-4}}$} & \textbf{0.417} & \textbf{0.79} \\
\bottomrule
\end{tabular}
\end{adjustbox}
\end{table}

\subsection{Impact of Stage-2 Tangent-Bundle Learning}
\label{sec:stage2_impact}

Stage-2 freezes the degradation encoder and introduces explicit manifold-velocity supervision. We observe a clear reduction in tangent violation~\ref{tab:stability}, indicating that the predicted manifold velocity aligns more closely with the tangent space of the learned manifold. This improves the geometric consistency of the learned degradation dynamics while preserving, and in some cases modestly improving restoration quality.

\subsection{Comparisons to Existing Restoration and Generative Baselines}
\label{sec:comparisons}

\paragraph{Distortion-based restoration comparison.}
Tables~\ref{tab:deblur_dpdd} and \ref{tab:real_resflow_style} compare our method against strong CNN-, transformer-, and flow-based baselines using PSNR, SSIM, and LPIPS. Across these benchmarks, our final hyperbolic model achieves competitive or improved distortion-based performance, with particularly clear gains on difficult deblurring examples and strong results on real-world degradations. The results shown for the state-of-the-art techniques are taken directly as reported by the authors. We have not used their checkpoints for evaluation as, for most of the recent works, the checkpoints and the model set-up are not available.

We do not claim a significant improvement of our model with respect to the state-of-the-art techniques. Our method have shown competitive results using manifold-based formulation. However, modeling the degradation as points on a manifold explicitly instead of a Euclidean embedding is more intuitive. Prior restoration methods treat degradation as conditioning tokens, degradation embeddings or implicit latent variables. Explicit manifold geometrical representation of degradation is aligned with the real world scenario, and our approach follows an intuitive way of learning without loss of performance.

\paragraph{Distributional quality comparison.}
Table~\ref{tab:diffusion_fid} reports FID/KID against diffusion-based baselines on task-specific perceptual benchmarks. Our method achieves competitive distributional quality while maintaining a restoration-oriented flow formulation, indicating that the proposed geometric conditioning improves restoration behavior without sacrificing perceptual realism.

\begin{table*}[t]
\centering
\scriptsize
\setlength{\tabcolsep}{2.2pt}
\renewcommand{\arraystretch}{1.03}
\caption{Comparison with diffusion-based methods using FID$\downarrow$ and KID$\downarrow$.
Deblur on RealBlur-J~\cite{rim2020realblur}, Dehaze on RESIDE~\cite{li2018reside}, 
Derain on RealRain~\cite{yang2020realrain}, and Desnow on Dense-Snow~\cite{chen2021densely}.}
\label{tab:diffusion_fid}
\begin{adjustbox}{max width=\linewidth}
\begin{tabular}{lcccccccc}
\toprule
Method &
\multicolumn{2}{c}{Deblur} &
\multicolumn{2}{c}{Dehaze} &
\multicolumn{2}{c}{Derain} &
\multicolumn{2}{c}{Desnow} \\
\cmidrule(lr){2-3}\cmidrule(lr){4-5}\cmidrule(lr){6-7}\cmidrule(lr){8-9}
& FID$\downarrow$ & KID$\downarrow$ & FID$\downarrow$ & KID$\downarrow$ & FID$\downarrow$ & KID$\downarrow$ & FID$\downarrow$ & KID$\downarrow$ \\
\midrule
Stable Diffusion~\cite{rombach2022ldm}              & 59.21 & 18.96 & 48.89 & 24.47 & 51.78 & 17.69 & 35.24 & 7.88 \\
InstructPix2Pix~\cite{brooks2023instructpix2pix}   & 57.38 & 19.37 & \textbf{33.48} & \textbf{12.76} & 54.12 & 17.87 & 42.01 & 8.54 \\
ControlNet~\cite{zhang2023controlnet}              & 52.30 & 17.19 & 37.02 & 15.45 & 52.55 & 15.22 & 34.36 & 5.70 \\
Diff-Plugin~\cite{diffplugin}                      & 51.81 & 14.63 & 34.68 & 14.38 & 50.55 & 13.41 & 34.30 & 5.20 \\
\midrule
Ours (Euclidean)                                   & 51.90 & 15.56 & 34.59 & 13.69 & 50.93 & 13.84 & 34.85 & 5.61 \\
Ours (Spherical)                                   & 50.13 & 13.91 & 33.68 & 13.01 & 48.83 & 12.87 & 33.14 & 4.27 \\
Ours (Hyperbolic)                                  & \textbf{50.05} & \textbf{13.67} & 33.95 & 13.17 & \textbf{48.09} & \textbf{12.84} & \textbf{33.02} & \textbf{4.13} \\
\bottomrule
\end{tabular}
\end{adjustbox}
\vspace{-1mm}
\end{table*}
\subsection{Ablations}
\label{sec:ablations}

We study the contribution of the proposed design along four axes: geometry choice, two-stage learning, conditioning strategy, and hyperparameter sensitivity.

\paragraph{Geometry choice.}
We compare Euclidean, spherical, and hyperbolic degradation manifolds. This ablation directly evaluates the effect of non-Euclidean structure on both restoration quality and geometric diagnostics.

\paragraph{Impact of Stage-2.}
We compare Stage-1-only training against the full two-stage formulation. This isolates the contribution of tangent-bundle supervision beyond learning a geometry-constrained degradation representation.

\paragraph{Conditioning strategy.}
We compare multiple ways of injecting degradation information into the restoration backbone, including direct feature modulation and cross-attention conditioning. Cross-attention performs best in our setting, which we attribute to its compatibility with the pretrained Stable Diffusion backbone and its ability to inject degradation information without disrupting latent feature structure.

\paragraph{Manifold dimension and curvature.}
We vary the manifold dimensionality and, for hyperbolic geometry, the curvature parameter. Very small dimensions can under-represent degradation variation, whereas larger dimensions provide diminishing returns and reduce interpretability. Similarly, small curvature approaches Euclidean behavior, while overly large curvature amplifies distance growth and can make optimization numerically sensitive. In practice, a moderate dimension and moderate curvature provide the best trade-off between expressivity and stability.

\begin{table}[t]
\centering
\caption{\textbf{Ablations} on DPDD (test). P/S/L: PSNR$\uparrow$/SSIM$\uparrow$/LPIPS$\downarrow$. D: diagnostic (constraint residual or severity correlation).}
\label{tab:ablation_horizontal}
\scriptsize
\setlength{\tabcolsep}{3pt}
\renewcommand{\arraystretch}{1.05}
\begin{adjustbox}{max width=\linewidth}
\begin{tabular}{@{}ccc@{}}
\toprule
\begin{tabular}{@{}lcccc@{}}
\multicolumn{5}{c}{\textbf{Geometry}}\\
\midrule
Setting & P & S & L & D\\
\midrule
Euclid & 25.43 & 0.808 & 0.174 & --\\
Sphere & 26.05 & 0.810 & 0.159 & $10^{-3}\!\downarrow$\\
Hyp    & \textbf{26.56} & \textbf{0.840} & \textbf{0.130} & $10^{-4}\!\downarrow$\\
\end{tabular}
&
\begin{tabular}{@{}lcccc@{}}
\multicolumn{5}{c}{\textbf{Dim.} $d_h$ (Hyp, S2)}\\
\midrule
Setting & P & S & L & D\\
\midrule
4  & 24.90 & 0.771 & 0.178 & 0.75$\uparrow$\\
8  & \textbf{26.56} & \textbf{0.840} & \textbf{0.130} & \textbf{0.79}$\uparrow$\\
16 & 25.53 & 0.804 & 0.159 & 0.76$\uparrow$\\
\end{tabular}
&
\begin{tabular}{@{}lcccc@{}}
\multicolumn{5}{c}{\textbf{Curv.} $c$ (Hyp, S2)}\\
\midrule
Setting & P & S & L & D\\
\midrule
0.1 & 25.38 & 0.794 & 0.172 & $10^{-3}\!\downarrow$\\
1.0 & \textbf{26.56} & \textbf{0.840} & \textbf{0.130} & \textbf{$10^{-4}\!\downarrow$}\\
5.0 & 24.14 & 0.772 & 0.179 & $10^{-3}\!\downarrow$\\
\end{tabular}
\\
\midrule
\multicolumn{3}{@{}l@{}}{
\begin{tabular}{@{}lcccc@{}}
\textbf{Two-stage / Conditioning (Hyperbolic)} & P & S & L & D\\
\midrule
Stage-1 only                   & 24.85 & 0.782 & 0.178 & $10^{-3}\!\downarrow$\\
Stage-1 + Stage-2              & \textbf{26.56} & \textbf{0.840} & \textbf{0.130} & \textbf{$10^{-4}\!\downarrow$}\\
AdaGN modulation (Stage-2)      & 24.89 & 0.776 & 0.174 & --\\
Cross-attention (Stage-2)       & \textbf{26.56} & \textbf{0.840} & \textbf{0.130} & --\\
\end{tabular}
}\\
\bottomrule
\end{tabular}
\end{adjustbox}
\vspace{0.3mm}
\scriptsize D is constraint residual (sphere: $|\|m\|_2-1|$, hyp: $|\langle m,m\rangle_L+1/c|$) or severity correlation (Spearman).
\end{table}

%% file: sec/conclusion.tex
\section{Conclusion}
\label{sec:conclusion}

We introduced a geometry-aware formulation for blind image restoration that models degradations as states on a structured manifold and learns restoration dynamics via latent flow matching in the joint space of image latents and degradation states. By replacing unconstrained Euclidean degradation embeddings with a degradation manifold, the method leverages geometric structure while retaining the flexibility of latent flow-based inference.

We further proposed a two-stage training strategy that first learns a geometry-constrained degradation representation jointly with the latent restoration flow, and then enforces tangent-bundle consistency for manifold dynamics. Experiments across multiple restoration tasks, including deblurring and adverse weather removal, show strong restoration quality alongside stable geometric behavior of the learned degradation states.

Finally, our analysis indicates that non-Euclidean geometries—particularly hyperbolic manifolds—offer a useful inductive bias, producing interpretable severity organization and consistent restoration trajectories. Overall, these findings suggest that injecting geometric structure into restoration representations can improve interpretability and robustness, and we hope they encourage broader exploration of geometry-aware latent generative restoration models.

%% file: sec/supplementary.tex
\clearpage
\setcounter{page}{1}
\maketitlesupplementary

\section{Overview}
In this paper, we perform image restoration by learning the degradation process and applying the reverse flow technique. However, unlike the existing flow-based techniques, we do not assume linear degradation geometry and rely on Euclidean interpolation. We model degradations as points on the manifold and formulate restoration as geodesic transport on the joint space.

\subsubsection{Problem Definition}
Let $x \in \mathbb{R}^{H\times W\times C}$ denote a clean image, and $y$, a degraded observation produced by an unknown corruption process $y = m(x)$. We aim to explicitly model degradations as elements of a continuous geometric space $m\in \mathcal{M}$ and perform restoration by transporting these degraded observations to a clean state along geometrically meaningful paths.

\subsection{Degradation Manifold Model}
We model degradations as elements of a smooth Riemannian manifold $(\mathcal{M}, g)$ of dimension $k << HWC$, where:
\begin{itemize}
    \item each point $m\in \mathcal{M}$ represents a degradation state,
    \item the Riemannian metric $g_m$ measures local degradation similarity,
    \item geodesics represent minimal transitions between degradation states.
\end{itemize}
The clean image corresponds to a canonical identity element $m_0\in \mathcal{M}$.

\paragraph{\normalfont \textbf{Assumption 1 (Manifold regularity):}} $\mathcal{M}$ is connected, geodesically complete, and admits a smooth exponential map $exp_m : T_m\mathcal{M}\rightarrow \mathcal{M}$.

Therefore, we define the joint state of image-manifold as:
\begin{equation}
    z = (x, m)\in \mathbb{R}^{H\times W\times C} \times \mathcal{M}
\end{equation}
Restoration is formulated as transport from $(x_1,m_1)$ to $(x_0,m_0)$ in the product space
\begin{equation}
    Z = \mathbb{R}^{H\times W\times C} \times \mathcal{M}
\end{equation}
This decouples image content from degradation geometry and allows mixed degradations to be interpreted as intermediate manifold states.

\subsubsection{Geodesic Flow Matching}
The restoration dynamics are defined using a time-dependent vector field
\begin{equation}
    v_{\theta} = (v_x(x,m,t), v_m(x,m,t))
\end{equation}
where, $v_{\theta}$ lies in Euclidean image space while $v_m(x,m, t)\in T_m\mathcal{M}$ lies in the tangent space of the manifold. The joint space is governed by
\begin{align}
    \frac{dx(t)}{dt} = v_x(x(t),m(t),t), \quad \frac{dm(t)}{dt} = v_m(x(t),m(t),t)
\end{align}
where, $\frac{dm(t)}{dt}$ is interpreted as a manifold-valued ordinary differential equation (ODE) on a random process
$\{z_t | 0 \leq t \leq 1\}$.

Given paired endpoints $(x_0, m_0)$ and $(x_1, m_1)$, we define geodesic interpolation on the manifold as:
\begin{equation}
    m_t = exp_{m_0} (t \mathrm{log}_{m_0}(m_1))
\end{equation}
where, $\mathrm{log}_{m_0}$ is the inverse exponential map. Therefore, the target velocity field is defined as:
\begin{equation}
    v^{\star}(z_t) = (x_1 - x_0, \dot{m_t}), \quad \dot{m_t} = \frac{d}{dt}m_t \in Tm_t\mathcal{M}
\end{equation}
where, $z_t$ defines the joint interpolation path given by:
\begin{equation}
    z_t = (x_t, m_t), \quad x_t = (1-t)x_0 + tx_1
\end{equation}

We define the Geodesic flow matching loss function with respect to the Euclidean Manifold space:
\begin{equation}
    \mathcal{L_{EM}} = \mathbb{E}_{(z_0, z_1), t}\|v_x(z_t, t)-(x_1-x_0)\|^2]+{\|v_m(z_t, t)-m_t\|}_{g_{m_t}}^2
\end{equation}
This objective reduces to Euclidean flow matching when $\mathcal{M}$ is flat.

\subsection{Proof of Convergence}
\label{sec:proof}
The following derivation proves how the optimization of the loss function mathematically guarantees the convergence of the generated distribution to the target data distribution on the manifold.

Let $\mathcal{M}$ be a smooth and orientable Riemannian manifold with a metric $g$ such that at any particular point $x\in\mathcal{M}$, $g:T_x\mathcal{M}\times T_x\mathcal{M}\to\mathbb{R}$ which is an inner product in the tangent space $T_x\mathcal{M}$ at point $x\in\mathcal{M}$. Furthermore, if $a$ is a vector lying over the tangent space $T_x\mathcal{M}$ then the length of the vector is defined as, $\lvert\rvert a\lvert\rvert_{g}=g(a,a)=\langle a,a\rangle_{x}$. Where $\langle\cdot\rangle_{x}$ is an inner product defined on the tangent space $T_x\mathcal{M}$.

Let, the distribution of the degraded images is defined as $p_0$ where $z_0 \sim p_0$ is a random degraded image embedding. And the distribution of clean images is defined as $p_1$ where $z_1 \sim p_1$ is a random clean image embedding. We assume that the feature embedding lies over the manifold $\mathcal{M}$. Where we define a target flow $z_t$ which is the ground truth probability path $p_t$ generated by a target vector field $v_t(t, x)$, as,
\begin{equation}
\frac{d}{dt}z_t = v_t(t,z)
\end{equation}

Now, as per the flow matching principle, the proposed method employs a neural network parameterized by $\theta$ estimating the flow $\hat{z}_t$ and generating a probability path $q_t$. The learned flow is defined as,
\begin{equation}
    \frac{d}{dt}\hat{z}_t = v_\theta(t, z)
\end{equation}
We define a family of curves $\gamma(s,t):[0,1]\times[0,1]\to \mathcal{M}$ such that $\gamma(0,t) = z_t$ and $\gamma(1,t) = \hat{z}_t$. Let $\delta(t)$ be the Riemannian distance $d_{\mathcal{M}}(\hat{z}_t, z_t)$ between the two flows $\hat{z}_t$ and $z_t$ at time $t$. So, $\delta(t)$ can be defined as the arc length of the curve between those points.
\begin{equation}
    \delta(t) = \int_0^1\lvert\rvert\frac{\partial\gamma(s,t)}{\partial s}\lvert\rvert_{g}ds=\int_0^1g\left(\frac{\partial\gamma(s,t)}{\partial s}, \frac{\partial\gamma(s,t)}{\partial s}\right) ds
\end{equation}
This $\delta(t)$ at any time point $t$ is denoting the prediction error of the neural network between true $z_t$ and the prediction $\hat{z}_t$.

Now, as per the first variation of arc length,
\begin{equation}
    \frac{d\delta(t)}{dt}=\Big\langle\frac{\partial\gamma(1,t)}{\partial s}, \frac{\partial\gamma(1,t)}{\partial t}\Big\rangle_{\hat{z}_t}-\Big\langle\frac{\partial\gamma(0,t)}{\partial s}, \frac{\partial\gamma(0,t)}{\partial t}\Big\rangle_{z_t}
    \label{eqn:inner product difference}
\end{equation}
We now have two inner products in two different tangent spaces ($T_{\hat{z}_t}\mathcal{M}$ and $T_{z_t}\mathcal{M}$). Hence, we can't take the difference between those two inner products. We use parallel transport to transport one tangent space to the other along the geodesic. Let $P_{z_t\to \hat{z}_t}$ denote the parallel transport operator along the geodesic from $z_t$ to $\hat{z}_t$. Now, with the help of parallel transport, Equation~(\ref{eqn:inner product difference}) can be written as,
\begin{equation}
   \frac{d\delta(t)}{dt}=\Big\langle\frac{\partial\gamma(1,t)}{\partial s}, \frac{\partial\gamma(1,t)}{\partial t}\Big\rangle_{\hat{z}_t}-\Big\langle P_{z_t\to \hat{z}_t}\frac{\partial\gamma(0,t)}{\partial s}, P_{z_t\to \hat{z}_t}\frac{\partial\gamma(0,t)}{\partial t}\Big\rangle_{\hat{z}_t} 
\end{equation}

Since, $\mathcal{M}$ is orientable hence, it has a unique Levi-Civita connection that preserves metric and torsion-free. As a result, moving along geodesic does not change length and orientation of a vector field. Therefore, $P_{z_t\to \hat{z}_t}\frac{\partial\gamma(0,t)}{\partial s}=\frac{\partial\gamma(1,t)}{\partial s}$ resulting the following equation,
\begin{equation}
    \frac{d\delta(t)}{dt}=\Big\langle\frac{\partial\gamma(1,t)}{\partial s}, \frac{\partial\gamma(1,t)}{\partial t}-P_{z_t\to \hat{z}_t}\frac{\partial\gamma(0,t)}{\partial t}\Big\rangle_{z_t}
\end{equation}
Applying Cauchy-Schwarz inequality the following upper bound can be established,
\begin{equation}
    \frac{d\delta(t)}{dt}\leq\Big\lvert\Big\rvert\frac{\partial\gamma(1,t)}{\partial s}\Big\lvert\Big\rvert_g \Big\lvert\Big\rvert v_{\theta}(t,z)-P_{z_t\to \hat{z}_t}\frac{\partial\gamma(0,t)}{\partial t}\Big\lvert\Big\rvert_g
\end{equation}
Now, we ensure that the $\frac{\partial\gamma(0,t)}{\partial s}$ is a normalized vector field. And, due to Levi-Civita connection,
$\Big\lvert\Big\rvert\frac{\partial\gamma(1,t)}{\partial s}\Big\lvert\Big\lvert_g = 1$ (As vector field preserves its length in parallel transport). hence, the above inequality can be written as,
\begin{equation}
    \frac{d\delta(t)}{dt}\leq\lvert\rvert v_{\theta}(t,z)-P_{z_t\to \hat{z}_t}v_t(t,z)\lvert\rvert_g
\end{equation}

We add and subtract $P_{z_t\to \hat{z}_t} v_{\theta}(t,z)$ inside the norm and apply the triangle inequality:
\begin{equation}
\frac{d}{dt} \delta(t) \leq  \lvert\rvert v_{\theta}(t,z)-P_{z_t\to \hat{z}_t} v_{\theta}(t,z)\lvert\rvert_g + \lvert\rvert P_{z_t\to \hat{z}_t} v_{\theta}(t,z) -P_{z_t\to \hat{z}_t} v_t(t, z)\lvert\rvert_g
\end{equation}

We assume the learned vector field $v_\theta (\cdot)$ is $L$-Lipschitz continuous with respect to the Riemannian metric. This bounds $\lvert\rvert v_{\theta}(t,z)-P_{z_t\to \hat{z}_t} v_{\theta}(t,z)\lvert\rvert_g\leq L\delta(t)$. Furthermore, Levi-Civita connection preserves the Riemannian norm, therefore $\lvert\rvert P_{z_t\to \hat{z}_t} v_{\theta}(t,z) -P_{z_t\to \hat{z}_t} v_t(t, z)\lvert\rvert_g$ simplifies to $\lvert\rvert v_{\theta}(t,z) -v_t(t, z)\lvert\rvert_g$. Combining those we obtain the following inequality,
\begin{equation}
    \frac{d}{dt} \delta(t) \leq L\delta(t) + \lvert\rvert v_{\theta}(t,z) -v_t(t, z)\lvert\rvert_g
\end{equation}

Since $\delta(0) = 0$ (both trajectories start at the same noise point $z_0$), we apply Grönwall's inequality to integrate the total prediction loss from $t=0$ to $t=1$ as,
\begin{equation}
    \delta(1) \le \int_0^1 e^{L(1-t)}\lvert\rvert v_{\theta}(t,z) -v_t(t, z)\lvert\rvert_g dt
\label{eqn:inequality}
\end{equation}
It can be observed from the above inequality that, the total prediction loss converges over time $t$ with exponential decay $e^{-Lt}$.
To obtain the final error bound, we square both sides of Equation~(\ref{eqn:inequality}) and apply the Cauchy-Schwarz inequality resulting the following bound over the total loss,
\begin{equation}
\delta(1)^2 \leq \left( \int_0^1 e^{2L(1-t)} dt \right) \left( \int_0^1 \lvert\rvert v_{\theta}(t,z) -v_t(t, z)\lvert\rvert_g^2 dt \right)
\end{equation}
Finally, we take the expectation over the noise distribution $z_0 \sim p_0$ and the clean data distribution $z_1 \sim p_1$, in order to obtain the mean squared error,
\begin{equation}
    \mathbb{E}_{z_0 \sim p_0, z_1 \sim p_1} [\delta(1)^2]\leq O(L^2)\int_0^1 \mathbb{E}_{z_0 \sim p_0, z_1 \sim p_1} \lvert\rvert v_{\theta}(t,z) -v_t(t, z)\lvert\rvert_g^2 dt
\end{equation}
where $\mathbb{E}_{z_0 \sim p_0, z_1 \sim p_1}\int_0^1 e^{2L(1-t)} dt\leq O(L^2)$. Therefore, the rate of convergence of the Riemannian manifold based flow matching is determined by $O(L^2)$.

\subsection{Theoretical Insight: Curvature-Induced Regularization}
\label{sec:theory}

We provide intuition for why non-Euclidean geometries can improve optimization stability when learning degradation representations.

\paragraph{Constraint-Induced Structure.}
In Euclidean space, degradation embeddings $m_1 \in \mathbb{R}^d$ are unconstrained. The model is free to scale embeddings arbitrarily, potentially leading to unstable optimization or poorly structured severity progression. In contrast, spherical and hyperbolic manifolds impose geometric constraints:
\begin{equation}
    \|m\|_2 = 1 
    \quad \text{(sphere)},
    \qquad
    \langle m,m\rangle_L = -\frac{1}{c}
    \quad \text{(hyperboloid)}.
\end{equation}
These constraints restrict the embedding to a curved submanifold, preventing unbounded growth and implicitly regularizing representation scale.

\paragraph{Geodesic Smoothness.}
In Euclidean space, straight-line interpolation between two embeddings does not enforce any intrinsic structure. On curved manifolds, geodesics follow curvature-aware trajectories defined by exponential maps. This induces smooth intrinsic interpolation:
\begin{equation}
    m_t = \mathrm{Exp}_{m_1}(t\,\mathrm{Log}_{m_1}(m_0)).
\end{equation}
Because geodesics minimize intrinsic distance, the induced degradation progression tends to be smoother and more consistent with a monotonic severity ordering.

\paragraph{Curvature and Sensitivity.}
Curvature influences how distances grow with displacement. In hyperbolic space, distances expand exponentially with radial displacement, while spherical geometry enforces bounded support. These properties affect gradient propagation and representation separation. In particular:

\begin{itemize}
    \item \textbf{Spherical geometry} bounds embeddings, preventing drift and limiting representation magnitude.
    \item \textbf{Hyperbolic geometry} provides increased representational resolution for distant points while maintaining intrinsic constraints.
\end{itemize}

Thus, curvature modulates how local perturbations affect global distances, acting as a form of geometric regularization.

\paragraph{Tangent Bundle Alignment.}
In Stage-2, manifold velocities are constrained to lie in $T_{m_t}\mathcal{M}$. The projection
\begin{equation}
    \tilde{v} = \Pi_{T_{m_t}\mathcal{M}}(v)
\end{equation}
eliminates components orthogonal to the manifold, reducing extrinsic drift. This further stabilizes learning by restricting dynamics to valid intrinsic directions.

\paragraph{Interpretation.}
We emphasize that curvature is introduced as an inductive bias rather than assumed to reflect true physical degradation geometry. Empirically, we observe that non-Euclidean manifolds reduce constraint violations, improve tangent alignment, and yield more stable training behavior across restoration tasks.

\section{Additional Method Details}

\subsection{Geometry-Specific Formulations}
\label{sec:geometry_ops_full}

We now provide explicit formulations for the degradation manifold 
$\mathcal{M}$ under Euclidean, spherical, and hyperbolic geometries referenced by the main paper. 
Our constructions follow standard Riemannian geometry and hyperbolic embedding literature 
\cite{doCarmo1992Riemannian,ganea2018hyperbolic,nickel2017poincare}.

\subsubsection{Euclidean Geometry}

In the Euclidean case,
\begin{equation}
    \mathcal{M} = \mathbb{R}^d,
\end{equation}
with the standard inner product.

Geodesics are straight lines:
\begin{equation}
    \gamma(t; m_1, m_0)
    =
    (1-t)m_1 + t m_0.
\end{equation}

The geodesic velocity is constant:
\begin{equation}
    \dot{m}_t = m_0 - m_1.
\end{equation}

The tangent space is identical to the ambient space:
\begin{equation}
    T_{m_t}\mathbb{R}^d = \mathbb{R}^d,
\end{equation}
and no projection is required.

\subsubsection{Spherical Geometry}

We consider the unit sphere
\begin{equation}
    \mathcal{M} = \mathbb{S}^{d-1}
    =
    \{ x \in \mathbb{R}^d \mid \|x\|_2 = 1 \}.
\end{equation}

The Riemannian metric is induced from the ambient Euclidean space.

\paragraph{Tangent Space.}
The tangent space at $x \in \mathbb{S}^{d-1}$ is
\begin{equation}
    T_x \mathbb{S}^{d-1}
    =
    \{ v \in \mathbb{R}^d \mid x^\top v = 0 \}.
\end{equation}

Projection onto the tangent space is given by
\begin{equation}
    \Pi_{T_x \mathbb{S}^{d-1}}(v)
    =
    v - (x^\top v) x.
\end{equation}

\paragraph{Exponential and Logarithmic Maps.}
Let $u \in T_x \mathbb{S}^{d-1}$. The exponential map is
\begin{equation}
    \mathrm{Exp}_x(u)
    =
    \cos(\|u\|) x
    +
    \sin(\|u\|)
    \frac{u}{\|u\|}.
\end{equation}

Given two points $x,y \in \mathbb{S}^{d-1}$, the logarithmic map is
\begin{equation}
    \mathrm{Log}_x(y)
    =
    \frac{\theta}{\sin \theta}
    \left(
        y - \cos\theta \, x
    \right),
\end{equation}
where
\begin{equation}
    \theta = \arccos(x^\top y).
\end{equation}

Geodesic interpolation is defined as
\begin{equation}
    \gamma(t; m_1, m_0)
    =
    \mathrm{Exp}_{m_1}
    \big(
        t \, \mathrm{Log}_{m_1}(m_0)
    \big).
\end{equation}

This corresponds to spherical linear interpolation (SLERP) \cite{Shoemake1985Slerp}.

\subsubsection{Hyperbolic Geometry (Lorentz Model)}

We adopt the Lorentz (hyperboloid) model of hyperbolic space 
\cite{nickel2017poincare,ganea2018hyperbolic}.

Let curvature parameter $c>0$. The hyperboloid is defined as
\begin{equation}
    \mathcal{M} = \mathbb{H}_c^d
    =
    \left\{
        x \in \mathbb{R}^{d+1}
        \mid
        \langle x,x \rangle_L = -\frac{1}{c},
        \;
        x_0 > 0
    \right\},
\end{equation}
where the Lorentzian inner product is
\begin{equation}
    \langle x,y \rangle_L
    =
    -x_0 y_0 + \sum_{i=1}^{d} x_i y_i.
\end{equation}

\paragraph{Tangent Space.}
The tangent space at $x \in \mathbb{H}_c^d$ is
\begin{equation}
    T_x \mathbb{H}_c^d
    =
    \{ v \in \mathbb{R}^{d+1}
    \mid
    \langle x, v \rangle_L = 0
    \}.
\end{equation}

Projection onto the tangent space is
\begin{equation}
    \Pi_{T_x \mathbb{H}_c^d}(v)
    =
    v + c \, \langle x, v \rangle_L \, x.
\end{equation}

\paragraph{Exponential Map.}
For $u \in T_x \mathbb{H}_c^d$, define
\begin{equation}
    \|u\|_L = \sqrt{\langle u,u \rangle_L}.
\end{equation}
The exponential map is
\begin{equation}
    \mathrm{Exp}_x(u)
    =
    \cosh\!\left(\sqrt{c}\|u\|_L\right) x
    +
    \frac{\sinh\!\left(\sqrt{c}\|u\|_L\right)}
         {\sqrt{c}\|u\|_L}
    u.
\end{equation}

\paragraph{Logarithmic Map.}
For $x,y \in \mathbb{H}_c^d$,
\begin{equation}
    \mathrm{Log}_x(y)
    =
    \frac{\mathrm{arcosh}(-c \langle x,y \rangle_L)}
         {\sqrt{(-c \langle x,y \rangle_L)^2 - 1}}
    \left(
        y + c \langle x,y \rangle_L x
    \right).
\end{equation}

Geodesic interpolation is defined as
\begin{equation}
    \gamma(t; m_1, m_0)
    =
    \mathrm{Exp}_{m_1}
    \big(
        t \, \mathrm{Log}_{m_1}(m_0)
    \big).
\end{equation}

\subsubsection{Intrinsic Velocity Norms}

For non-Euclidean geometries, manifold losses use intrinsic norms:

\paragraph{Sphere.}
\begin{equation}
    \|v\|_{g(x)}^2 = \|v\|_2^2.
\end{equation}

\paragraph{Hyperboloid.}
\begin{equation}
    \|v\|_{g(x)}^2 = \langle v,v \rangle_L.
\end{equation}

This ensures that manifold velocity supervision is performed in the appropriate Riemannian metric.


\begin{algorithm}[t]
\caption{Inference for Manifold-Conditioned Latent Flow Matching}
\label{alg:inference}
\begin{algorithmic}[1]
\REQUIRE Degraded image $y$; frozen Stable Diffusion autoencoder $(E_\phi, D_\phi)$ \cite{rombach2022ldm}; frozen degradation encoder $g_\psi$; clean anchor $m_0 \in \mathcal{M}$; trained UNet velocity head $v_\theta^z$ (and optionally $v_\theta^m$); steps $K$; solver step schedule $\{t_k\}_{k=0}^{K}$ with $t_0=0$, $t_K=1$.
\ENSURE Restored image $\hat{x}$.
\vspace{1mm}
\STATE Encode latent: $z \leftarrow E_\phi(y)$.
\STATE Degradation state: $m_1 \leftarrow g_\psi(z)$.
\FOR{$k = 0$ to $K-1$}
    \STATE Sample (or set) time: $t \leftarrow t_k$.
    \STATE Manifold interpolation: $m_t \leftarrow \gamma(t; m_1, m_0)$ \hfill (geodesic on $\mathcal{M}$)
    \STATE Conditioning tokens: $\mathbf{c}_t \leftarrow [\mathrm{Emb}(\varnothing),\, W_m m_t]$.
    \STATE Predict latent velocity: $v \leftarrow v_\theta^z(z, t, \mathbf{c}_t)$.
    \STATE Update latent (Euler): $z \leftarrow z + (t_{k+1}-t_k)\, v$.
    \STATE \textit{(Optional, Stage-2 dynamics)} Predict manifold velocity $u \leftarrow v_\theta^m(z, m_t, t)$; project $u \leftarrow \Pi_{T_{m_t}\mathcal{M}}(u)$; update $m \leftarrow \mathrm{Exp}_{m_t}((t_{k+1}-t_k)u)$.
\ENDFOR
\STATE Decode: $\hat{x} \leftarrow D_\phi(z)$.
\RETURN $\hat{x}$.
\end{algorithmic}
\end{algorithm}

\subsection{Inference Solver}
\label{sec:supp_inference_solver}

At inference time, restoration is performed by integrating the learned latent vector field in the latent space of the frozen Stable Diffusion autoencoder. Given a degraded image $y$, we first encode it as
\begin{equation}
z_0 = E_\phi(y),
\end{equation}
and obtain the degradation embedding
\begin{equation}
m_1 = g_\psi(z_0),
\end{equation}
where $g_\psi$ is the frozen degradation encoder after Stage-1 training. The clean anchor $m_0$ is fixed by the chosen manifold parameterization, and the conditioning state at time $t$ is defined by the geodesic interpolation
\begin{equation}
m_t = \gamma(t; m_1, m_0).
\end{equation}
The manifold state is projected to the UNet context dimension and concatenated with the empty-text token to form the conditioning tokens used in cross-attention.

We integrate the latent dynamics
\begin{equation}
\frac{dz}{dt} = v_\theta^z(z_t, t, m_t)
\end{equation}
using a fixed-step explicit solver. In the main experiments, we use Heun's method (RK2), which provides a favorable trade-off between stability and computational cost compared with single-step Euler updates.

For a step size $\Delta t = t_{k+1}-t_k$, Heun integration proceeds as
\begin{align}
k_1 &= v_\theta^z(z_k, t_k, m_{t_k}), \\
\tilde z_{k+1} &= z_k + \Delta t \, k_1, \\
k_2 &= v_\theta^z(\tilde z_{k+1}, t_{k+1}, m_{t_{k+1}}), \\
z_{k+1} &= z_k + \frac{\Delta t}{2}(k_1 + k_2).
\end{align}
After the final step, the restored latent is decoded by the frozen decoder:
\begin{equation}
\hat x = D_\phi(z_K).
\end{equation}

In our main implementation, the manifold trajectory is used as a conditioning signal during inference, i.e., $m_t$ is recomputed from the geodesic path $\gamma(t;m_1,m_0)$ at each integration step. Although Stage-2 additionally learns a manifold vector field $v_\theta^m$, we found conditioning-only inference to be more stable and sufficient for the main restoration results. This also keeps the inference procedure simple and avoids compounding numerical errors from simultaneously integrating latent and manifold states. 

\paragraph{Why Heun/RK2.}
We prefer Heun over Euler because the restoration field is conditioned on both latent content and a time-dependent manifold state, which can make the dynamics locally stiff under severe degradations. In this setting, the predictor--corrector update of Heun consistently produces cleaner reconstructions at the same number of steps. Euler remains usable for debugging and ablation, but is slightly less stable in our experiments.

\subsection{Numerical Stability Implementation Details}
\label{sec:supp_numerical_stability}

To train geometry-aware latent flow models we found several implementation choices necessary in practice to ensure stable optimization and numerically valid manifold operations. We summarize the main stability practices below.


\paragraph{(1) Stable hyperbolic projection.}
For hyperbolic geometry, points are represented on the upper sheet of the Lorentz hyperboloid,
\begin{equation}
\langle x,x\rangle_L = -\frac{1}{c}, \qquad x_0>0.
\end{equation}
Given spatial coordinates $x_s$, we recover the time-like coordinate as
\begin{equation}
x_0 = \sqrt{\frac{1}{c} + \|x_s\|_2^2},
\end{equation}
with a small lower clamp inside the square root for numerical safety. This prevents invalid hyperboloid states and ensures all embeddings remain on the upper sheet.


\paragraph{(2) Clamps for inverse trigonometric and inverse hyperbolic functions.}
Both spherical and hyperbolic logarithmic maps require inverse nonlinearities whose domains can be violated by small floating-point errors. We therefore clamp arguments before evaluation:
\begin{itemize}
    \item For spherical geometry, inputs to $\arccos$ are clamped to $[-1+\delta, 1-\delta]$.
    \item For hyperbolic geometry, inputs to $\mathrm{arcosh}$ are clamped to values strictly larger than $1$, i.e., $1+\delta$.
\end{itemize}
Without these clamps, numerical noise can produce NaNs during training or analysis.


\paragraph{(3) Avoiding dead-start initialization for hyperbolic encoders.}
A naive zero initialization of the direction head in the hyperbolic degradation encoder causes the tangent direction to collapse to zero, leading to a dead-start where all embeddings remain near the origin with negligible gradients. To prevent this, we initialize the direction head with small non-zero weights while keeping the radius head bounded. This allows the encoder to explore meaningful tangent directions from the beginning of training.

\paragraph{(4) Bounded radius parameterization for hyperbolic embeddings.}
To prevent uncontrolled growth of tangent vectors in the hyperbolic encoder, we parameterize the tangent radius using a bounded nonlinearity, e.g.,
\begin{equation}
r = r_{\max}\tanh(r_{\mathrm{raw}}),
\end{equation}
or an equivalent positive bounded variant. The final tangent vector is then written as
\begin{equation}
u = r \, \frac{u_{\mathrm{raw}}}{\|u_{\mathrm{raw}}\|_2 + \varepsilon}.
\end{equation}
This makes the severity magnitude interpretable and prevents large excursions that can destabilize the exponential map.

\paragraph{(5) Freezing the degradation encoder in Stage-2.}
Jointly learning the degradation encoder and manifold velocity supervision can create a moving-target problem: the manifold state $m_t$ changes while the model is simultaneously asked to regress its derivative. Freezing the degradation encoder in Stage-2 removes this instability and leads to a much cleaner tangent-bundle supervision signal.




\section{Additional Experimental Setup}
\label{sec:supp_setup}

\subsection{Training Datasets and Protocols}
\label{sec:supp_datasets}

We train task-specific models under the standard protocol associated with each benchmark.
For deblurring, we use both DPDD~\cite{abdelhamed2019dpdd} and GoPro~\cite{nah2017deepdeblur}, training separate models on each benchmark.
DPDD is used for defocus deblurring experiments, while GoPro is used for dynamic-scene motion deblurring.
For dehazing, we use the RESIDE OTS training split~\cite{li2018reside}, which is the standard large-scale synthetic training benchmark for single-image dehazing.
For desnowing, we train on Snow100K~\cite{liu2018desnownet}, following the standard paired synthetic training protocol introduced with DesnowNet.
For deraining, we follow the merged synthetic deraining training protocol commonly used in MPRNet~\cite{zamir2021mprnet}, corresponding to the mixed-training Rain13K benchmark setting summarized in the unified deraining benchmark of Chen \emph{et al.}~\cite{chen2023survey}.
Across all tasks, we use a crop size of $512\times512$ during training.

\subsection{Training Details}
\label{sec:supp_training_details}

Unless otherwise stated, both Stage-1 and Stage-2 are optimized using AdamW with initial learning rate $1\times 10^{-5}$ and weight decay $1\times 10^{-2}$.
The degradation encoder is trainable in Stage-1 and frozen in Stage-2.

Because the four restoration tasks differ substantially in dataset size and convergence speed, we do not fix a single universal epoch count across all tasks.
Instead, Stage-1 is trained until the latent flow-matching objective stabilizes on the validation split, after which Stage-2 is initialized from the Stage-1 checkpoint and trained for at least 50 additional epochs, with longer training for larger datasets when needed.
In all cases, the same two-stage protocol is used:
Stage-1 learns a geometry-constrained degradation representation jointly with latent restoration flow, and Stage-2 freezes the degradation encoder and learns tangent-bundle-aligned manifold dynamics while keeping the latent restoration objective active.

\subsection{Inference Details}
\label{sec:supp_inference_details}

At inference, restoration is performed by integrating the latent vector field in the latent space of the frozen Stable Diffusion autoencoder.
We use the fixed-step Heun/RK2 solver described in Sec.~\ref{sec:supp_inference_solver}.
For all tasks, we use a uniform step schedule with 5 inference steps.
The same solver configuration is used across all geometries within a task to ensure fair comparison.

\section{Additional Quantitative Results}
\label{sec:supp_additional_quant}

This section extends the empirical analysis of the main paper along three directions.
First, we evaluate the proposed hyperbolic model on additional deblurring benchmarks beyond the DPDD setting used in the main paper.
Second, we provide broader quantitative evidence for restoration under synthetic weather degradations.
Third, we include supplementary studies on inference efficiency, ablations, and cross-dataset generalization to further characterize the behavior of the proposed geometry-aware latent flow.

\subsection{Additional Deblurring Results on GoPro and RealBlur}
\label{sec:supp_gopro_realblur}

To assess the proposed method beyond the defocus-blur setting of DPDD, we train the final hyperbolic model on GoPro~\cite{nah2017deepdeblur} and evaluate it on GoPro, RealBlur-J, and RealBlur-R.
This experiment serves two purposes: it tests the method on a standard synthetic motion deblurring benchmark and provides an initial measure of transfer to real blurry images.

Table~\ref{tab:deblur_gopro_realblur} reports the quantitative results.
When trained on GoPro, the proposed hyperbolic model achieves strong restoration performance on the GoPro test set while remaining competitive on the real-world RealBlur-J and RealBlur-R benchmarks.
These results indicate that the geometry-aware latent flow is not limited to the defocus-blur regime of DPDD, but also remains effective in dynamic-scene motion deblurring and transfers reasonably to real blur distributions.

\begin{table*}[t]
\centering
\caption{\textbf{Comprehensive deblurring comparison on synthetic and real benchmarks.}
We report PSNR$\uparrow$, SSIM$\uparrow$, LPIPS$\downarrow$, and FID$\downarrow$ on GoPro, and PSNR$\uparrow$/SSIM$\uparrow$ on RealBlur-J and RealBlur-R.
Best and second-best results are highlighted in \textbf{bold} and \underline{underline}, respectively.}
\label{tab:deblur_gopro_realblur}
\small
\setlength{\tabcolsep}{4pt}
\renewcommand{\arraystretch}{1.12}
\begin{adjustbox}{max width=\linewidth}
\begin{tabular}{lcccccccc}
\toprule
\multirow{2}{*}{Method} & \multicolumn{4}{c}{GoPro} & \multicolumn{2}{c}{RealBlur-J} & \multicolumn{2}{c}{RealBlur-R} \\
\cmidrule(lr){2-5} \cmidrule(lr){6-7} \cmidrule(lr){8-9}
& PSNR$\uparrow$ & SSIM$\uparrow$ & LPIPS$\downarrow$ & FID$\downarrow$ & PSNR$\uparrow$ & SSIM$\uparrow$ & PSNR$\uparrow$ & SSIM$\uparrow$ \\
\midrule
MPRNet~\cite{zamir2021mprnet} & 32.66 & 0.959 & 0.089 & 20.18 & 28.70 & 0.873 & 35.99 & 0.952 \\
FocalNet~\cite{cui2023focal} & 33.10 & 0.962 & -- & -- & -- & -- & -- & -- \\
Restormer~\cite{zamir2022restormer} & 32.92 & 0.961 & 0.084 & 19.33 & 28.96 & \underline{0.879} & \underline{36.19} & 0.957 \\
NAFNet~\cite{chen2022nafnet} & \textbf{33.71} & \textbf{0.966} & -- & -- & \underline{28.96} & 0.873 & 35.89 & \textbf{0.962} \\
DTPM~\cite{ye2024learning} & 32.09 & 0.932 & 0.08 & 10.02 & -- & -- & -- & -- \\
\midrule
Ours (Hyperbolic) & \underline{33.54} & \underline{0.965} & \textbf{0.074} & \textbf{8.36} & \textbf{29.97} & \textbf{0.903} & \textbf{36.27} & \underline{0.96} \\
\bottomrule
\end{tabular}
\end{adjustbox}
\end{table*}

\begin{figure*}[ht]
    \centering
    \includegraphics[width=1.0\linewidth]{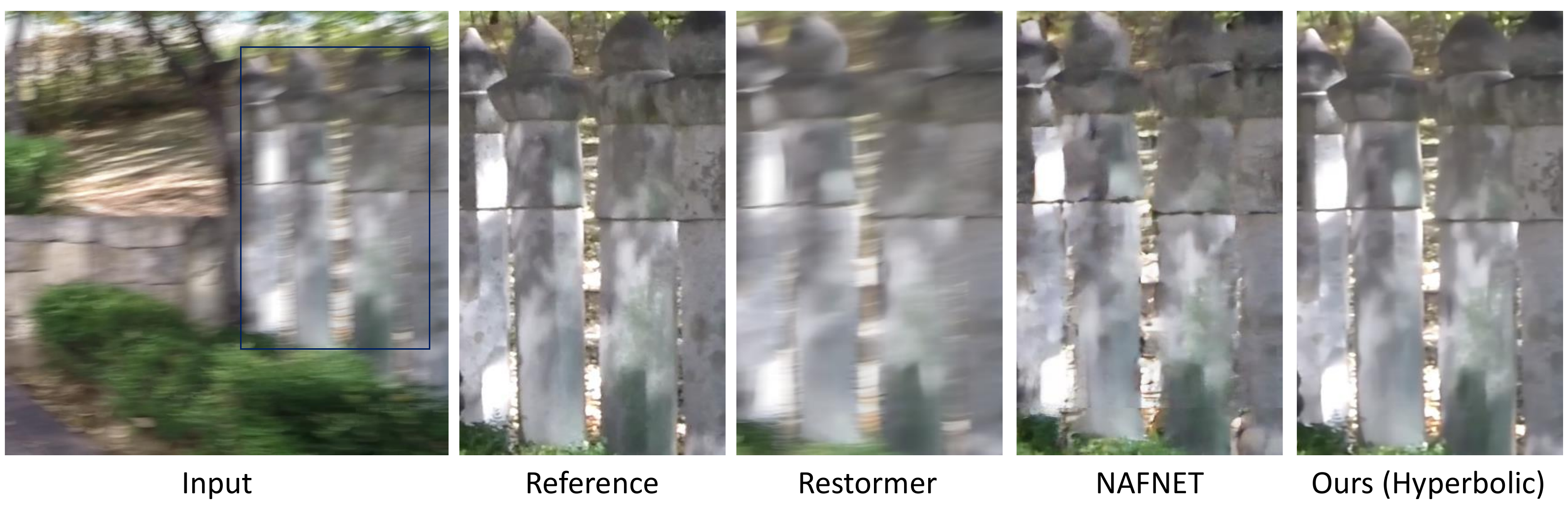}
    \caption{\textbf{Qualitative comparison on GoPro.}
Visual comparison of our final hyperbolic model against Restormer and NAFNet on representative GoPro motion deblurring examples.
Our method restores sharper edges, recovers fine textures more faithfully, and reduces residual motion blur and oversmoothing artifacts more effectively than the competing baselines.
Zoomed-in regions highlight improved reconstruction of thin structures and high-frequency details.}
\label{fig:gopro_qual}
\end{figure*}

\begin{figure*}[ht]
    \centering
    \includegraphics[width=1.0\linewidth]{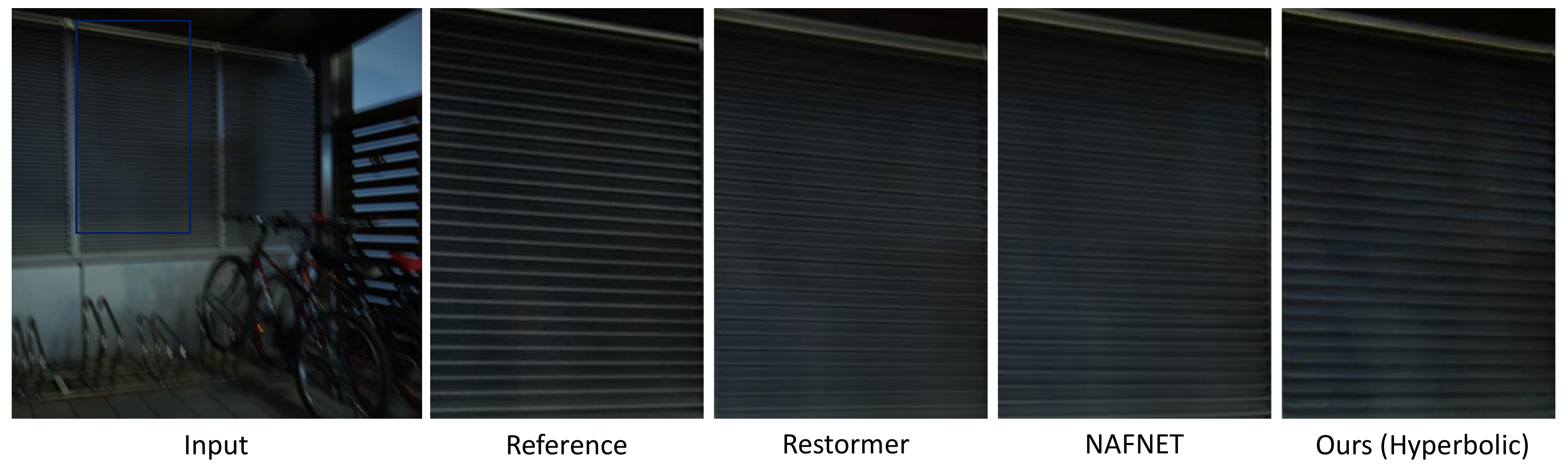}
    \caption{\textbf{Qualitative comparison on RealBlur-J.}
Visual comparison of our final hyperbolic model against Restormer and NAFNet on real blurry images from RealBlur-J.
Our method better restores image sharpness and local contrast while preserving natural appearance, yielding clearer structures and fewer oversmoothing artifacts than the baselines.
These examples indicate that the proposed geometry-aware latent flow generalizes effectively from GoPro training to real-world blur.}
\label{fig:realblurj_qual}
\end{figure*}

Figures~\ref{fig:gopro_qual} and \ref{fig:realblurj_qual} provide qualitative comparisons on GoPro and RealBlur-J, respectively.
Compared with strong restoration baselines such as Restormer and NAFNet, our method restores sharper structures, preserves fine details more faithfully, and reduces residual blur and oversmoothing artifacts more effectively.

\subsection{Cross-Dataset Generalization}
\label{sec:supp_blind_generalization}

While the main paper focuses on standard in-distribution evaluation, blind restoration also requires robustness to degradation distributions that differ from those seen during training.
To assess this more directly, we perform cross-dataset deblurring evaluation in which a model trained on one dataset is directly tested on another benchmark without retraining.

\begin{figure*}[ht]
    \centering
    \includegraphics[width=1.0\linewidth]{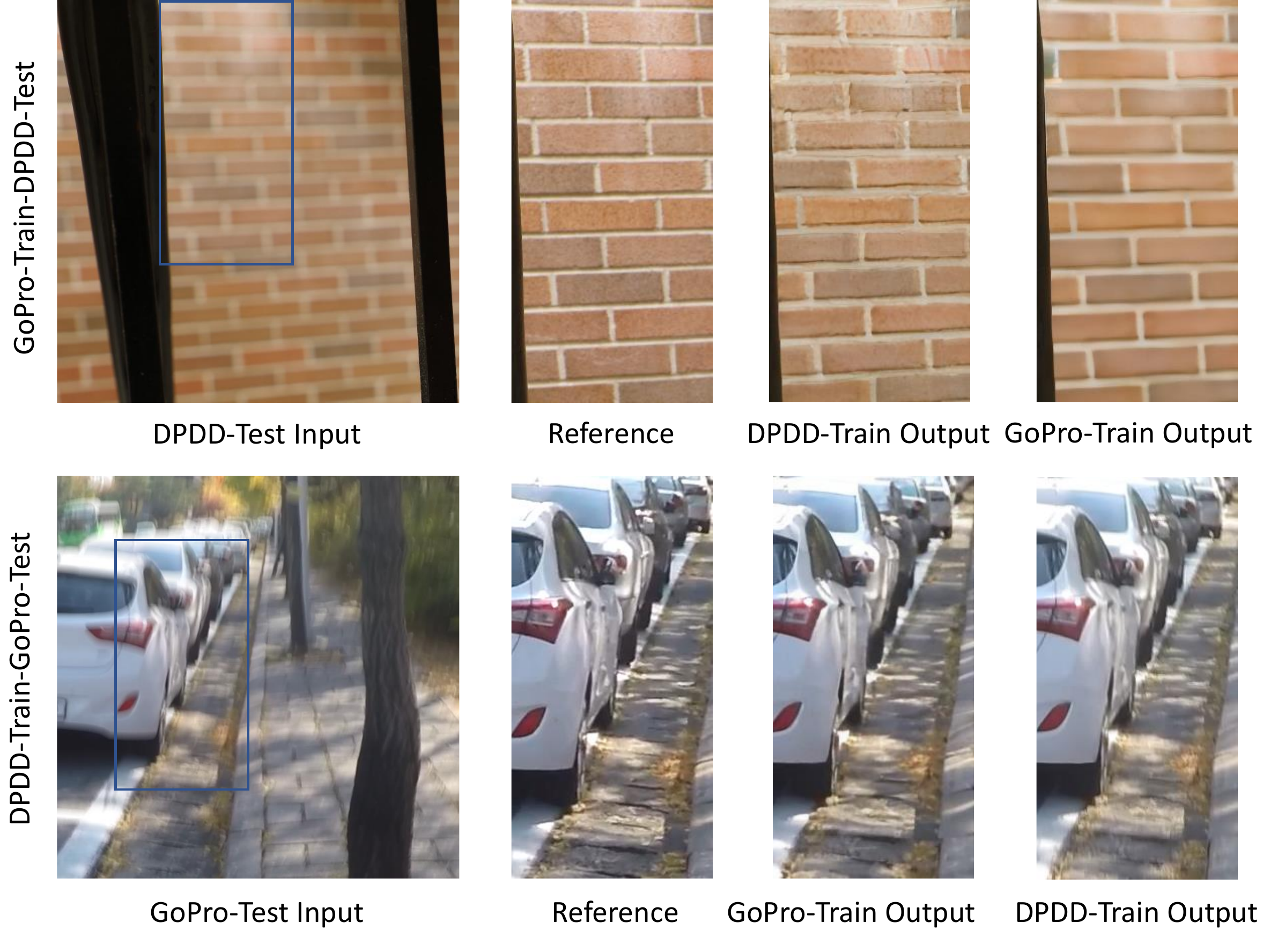}
    \caption{\textbf{Qualitative cross-dataset deblurring analysis.}
Top row: results of the model trained on GoPro and tested on a DPDD image.
Bottom row: results of the model trained on DPDD and tested on a GoPro image.
Although the blur characteristics differ substantially across the two datasets, the proposed hyperbolic model retains meaningful restoration capability under this cross-dataset setting, illustrating its ability to generalize beyond the training distribution.}
\label{fig:cross_dataset_qual}
\end{figure*}

Table~\ref{tab:cross_dataset_deblur} summarizes this setting for DPDD and GoPro. Because DPDD and GoPro differ substantially in blur type (defocus vs dynamic motion blur), this setting is intentionally challenging.
This protocol introduces both dataset shift and blur-distribution shift, and therefore provides a stronger test of blind restoration behavior than standard in-domain evaluation. Figure~\ref{fig:cross_dataset_qual} further illustrates this behavior qualitatively, showing that the proposed hyperbolic model retains meaningful restoration capability even when evaluated under cross-dataset blur transfer.
Our goal here is not to claim full domain invariance, but to examine whether the proposed geometry-aware degradation representation captures transferable structure beyond the training distribution.

As expected, performance decreases under cross-dataset transfer relative to in-domain evaluation.
However, the model retains meaningful restoration capability under this distribution shift, supporting the view that the learned manifold representation provides a useful inductive bias even when the blur statistics differ from training.

\begin{table}[t]
\centering
\caption{\textbf{Cross-dataset deblurring generalization.}
Models are trained on one dataset and directly evaluated on another without retraining.}
\label{tab:cross_dataset_deblur}
\small
\setlength{\tabcolsep}{4pt}
\renewcommand{\arraystretch}{1.10}
\begin{adjustbox}{max width=\linewidth}
\begin{tabular}{llccc}
\toprule
\textbf{Train} & \textbf{Test} & PSNR$\uparrow$ & SSIM$\uparrow$ & LPIPS$\downarrow$ \\
\midrule
DPDD  & DPDD  & 26.56 & 0.840 & 0.130 \\
DPDD  & GoPro & 32.01 & 0.932 & 0.091 \\
\midrule
GoPro & GoPro & 33.54 & 0.965 & 0.074 \\
GoPro & DPDD  & 25.98 & 0.829 & 0.151 \\
\bottomrule
\end{tabular}
\end{adjustbox}
\end{table}

\subsection{Synthetic Weather Restoration Benchmarks}
\label{sec:supp_synthetic_weather}

In addition to the real-world evaluations reported in the main paper, we include quantitative comparisons on synthetic weather-degradation benchmarks.
Specifically, we report PSNR, SSIM, and LPIPS on Snow100K~\cite{liu2018desnownet}, Outdoor-Rain~\cite{li2019heavy}, and Dense-Haze~\cite{ancuti2019dense} for desnowing, deraining, and dehazing, respectively.

These controlled paired benchmarks complement the real-world evaluations by providing a direct measurement of pixel-level restoration quality under standardized synthetic degradations.
The results show that the proposed hyperbolic model remains competitive across all three tasks, indicating that the geometry-aware latent flow formulation is effective not only on real-world restoration settings but also under standard synthetic evaluation protocols.

\begin{table*}[t]
\vspace{-2mm}
\centering
\caption{\textbf{Synthetic datasets.} Quantitative comparisons on Snow100K~\cite{liu2018desnownet},
Outdoor-Rain~\cite{li2019heavy}, and Dense-Haze~\cite{ancuti2019dense}. We report PSNR$\uparrow$, SSIM$\uparrow$, and LPIPS$\downarrow$.
\textemdash~denotes metrics not reported by the corresponding method.}
\label{tab:synthetic_tasks}
\setlength{\tabcolsep}{2.3pt}
\renewcommand{\arraystretch}{1.10}
\scriptsize
\begin{adjustbox}{max width=\linewidth}
\begin{tabular}{l|ccc|ccc|ccc}
\toprule
\multirow{2}{*}{Method} &
\multicolumn{3}{c|}{\textbf{Desnowing}} &
\multicolumn{3}{c|}{\textbf{Deraining}} &
\multicolumn{3}{c}{\textbf{Dehazing}} \\
\cmidrule(lr){2-4}\cmidrule(lr){5-7}\cmidrule(lr){8-10}
& PSNR$\uparrow$ & SSIM$\uparrow$ & LPIPS$\downarrow$
& PSNR$\uparrow$ & SSIM$\uparrow$ & LPIPS$\downarrow$
& PSNR$\uparrow$ & SSIM$\uparrow$ \\
\midrule
MPRNet~\cite{zamir2021mprnet}           & 29.76 & 0.894 & 0.049 & 28.03 & 0.919 & 0.089 & \textemdash & \textemdash & \\
NAFNet~\cite{chen2022nafnet}           & 30.06 & 0.901 & 0.051 & 29.59 & 0.902 & 0.085 & \textemdash & \textemdash & \\
Restormer~\cite{zamir2022restormer}     & 30.52 & 0.909 & 0.047 & 29.97 & 0.921 & 0.074 & \textemdash & \textemdash &  \\
DTPM-4~\cite{ye2024learning}            & 30.92 & \underline{0.917} & 0.034 & 30.99 & 0.934 & 0.063 & \textemdash & \textemdash & \\
ResFlow \cite{Qin2025}        & \textbf{31.86} & \textbf{0.917} & \textbf{0.030} & \textbf{32.82} & \textbf{0.936} & \textbf{0.051} & 17.12 & 0.59 & \\
\midrule
\textbf{Ours (Euclidean)}        & 30.45 & 0.890 & 0.032 & 30.78 & 0.926 & 0.057 & 18.72 & \underline{0.510} & \\
\textbf{Ours (Spherical)}        & 30.80 & 0.909 & 0.031 & 31.25 & 0.929 & 0.056 & \underline{18.56} & 0.509 & \\
\textbf{Ours (Hyperbolic)}        & \underline{31.01} & 0.911 & \underline{0.031} & \underline{32.09} &\underline{0.929} & \underline{0.056} & \textbf{18.58} & \textbf{0.514} & \\
\bottomrule
\end{tabular}
\end{adjustbox}
\vspace{-2mm}
\end{table*}

\subsection{Effect of Inference Step Count}
\label{sec:supp_inference_steps}

We further study the effect of the number of Heun/RK2 inference steps on restoration quality.
Table~\ref{tab:inference_steps} reports results using different step counts on the representative deblurring task.

\begin{figure*}[ht]
    \centering
    \includegraphics[width=1.0\linewidth]{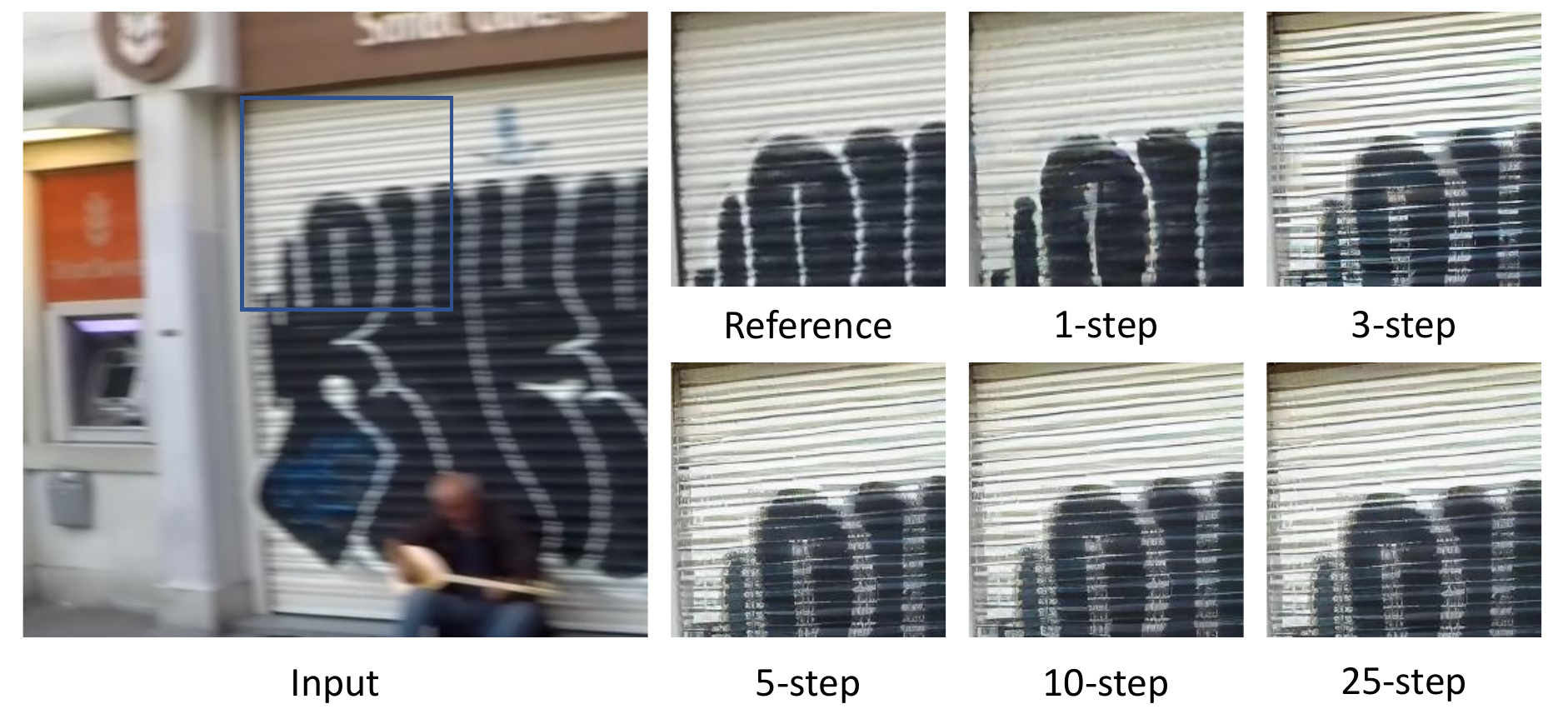}
    \caption{\textbf{Effect of inference step count on restoration quality.}
Qualitative deblurring results using 1, 3, 5, 10, and 25 Heun/RK2 inference steps.
Restoration quality improves noticeably from 1 to 5 steps, with sharper structures, clearer textures, and reduced residual blur.
Beyond 5 steps, the visual improvements become marginal, indicating that the proposed latent flow largely saturates after a small number of inference steps while remaining efficient at test time.}
\label{fig:multistep_inference}
\end{figure*}

We observe that restoration quality improves substantially when increasing the number of steps from very small values to the default 5-step setting, after which the gains become progressively smaller. This trend is also visible qualitatively in figure~\ref{fig:multistep_inference}.
This behavior indicates that the proposed latent flow reaches strong restoration quality with only a small number of solver steps, while additional steps mainly provide diminishing returns.
These results support the efficiency of the proposed formulation at test time.

\begin{table}[t]
\centering
\caption{\textbf{Effect of inference step count.}
Restoration quality on the representative deblurring task as the number of Heun/RK2 inference steps is varied.}
\label{tab:inference_steps}
\small
\setlength{\tabcolsep}{4pt}
\renewcommand{\arraystretch}{1.10}
\begin{adjustbox}{max width=\linewidth}
\begin{tabular}{lccccc}
\toprule
\textbf{Metric} & \textbf{1} & \textbf{3} & \textbf{5} & \textbf{10} & \textbf{25} \\
\midrule
PSNR$\uparrow$  & 32.17 & 33.21 & 33.54 & 33.59 & 33.54 \\
SSIM$\uparrow$  & 0.942 & 0.959 & 0.965 & 0.967 & 0.965 \\
LPIPS$\downarrow$ & 0.08 & 0.079 & 0.074 & 0.0.072 & 0.074 \\
\bottomrule
\end{tabular}
\end{adjustbox}
\end{table}

\section{Conclusion}
\label{sec:supp_conclusion}

This supplementary material provides additional technical and empirical support for the main paper.
We included full geometry operations, implementation and numerical-stability details, additional quantitative evaluations on deblurring and synthetic weather benchmarks, cross-dataset generalization experiments, extended ablations, and further qualitative and geometric diagnostics.

Taken together, these supplementary results reinforce the main conclusions of the paper.
First, the proposed geometry-aware latent flow extends effectively beyond the primary DPDD setting to additional deblurring benchmarks such as GoPro and RealBlur.
Second, the method remains competitive across synthetic dehazing, deraining, and desnowing benchmarks, showing that the proposed formulation is not tied to a single restoration regime.
Third, the additional ablations and diagnostics further support the role of non-Euclidean geometry, particularly hyperbolic geometry, in yielding structured degradation representations and stable restoration behavior.

Overall, the supplementary experiments strengthen the evidence that modeling degradation states on a Riemannian manifold provides a useful and interpretable inductive bias for blind image restoration.

\clearpage
